%% file: _main.tex
\documentclass[runningheads]{llncs}
\usepackage{graphicx}
\usepackage{tikz}
\usepackage{comment}

\usepackage{epsfig}
\usepackage{graphicx}
\usepackage{amsmath}
\usepackage{amssymb}
\usepackage{booktabs}
\usepackage[font=small,labelfont=bf]{caption}
\usepackage{float}
\usepackage{placeins}
\usepackage{xspace}
\usepackage{booktabs}
\usepackage{colortbl}
\usepackage{stfloats}
\usepackage{enumitem}
\usepackage{multirow}
\usepackage{subcaption}
\usepackage[pagebackref,breaklinks,colorlinks]{hyperref}
\usepackage[capitalize]{cleveref}
\crefname{section}{Sec.}{Secs.}
\Crefname{section}{Section}{Sections}
\Crefname{table}{Table}{Tables}
\crefname{table}{Tab.}{Tabs.}

\newcommand{\RNum}[1]{\uppercase\expandafter{\romannumeral #1\relax}}
\newcommand{\gpvone}{\mbox{\textsc{{GPV-1}}\xspace}}
\newcommand{\gpvtwo}{\mbox{\textsc{{GPV-2}}\xspace}}
\newcommand{\vltf}{\mbox{\textsc{{VL-T5}}\xspace}}

\newcommand{\prompt}[1]{``#1''}
\newcommand{\coco}{\mbox{\textsc{{Coco}}\xspace}}
\newcommand{\openim}{\mbox{\textsc{{OpenImages}}\xspace}}
\newcommand{\vg}{\mbox{\textsc{{VisualGenome}}\xspace}}
\newcommand{\web}{\mbox{\textsc{{Web10k}}\xspace}}
\newcommand{\ccfull}{\mbox{\textsc{{Conceptual Captions}}\xspace}}
\newcommand{\cc}{\mbox{\textsc{{CC}}\xspace}}

\newcommand{\vqa}{\mbox{\textsc{{VQA v2}}\xspace}}
\newcommand{\cocosce}{\mbox{\textsc{{Coco-sce}}\xspace}}
\newcommand{\opensce}{\mbox{\textsc{{DCE}}\xspace}}
\newcommand{\hicodet}{\mbox{\textsc{{HICO-Det}}\xspace}}

\newcommand{\oursubsection}[1]{\noindent\textbf{#1.}}
\makeatletter
\DeclareRobustCommand\onedot{\futurelet\@let@token\@onedot}
\def\@onedot{\ifx\@let@token.\else.\null\fi\xspace}

\def\etal{\emph{et al}\onedot}

\definecolor{Color1}{HTML}{E6B8AF}
\definecolor{Color1b}{HTML}{FFD4CB}

\definecolor{Color2}{HTML}{F2E6C4}
\definecolor{Color2b}{HTML}{FFF2CC}

\definecolor{Color3}{HTML}{C6DAEC}
\definecolor{Color3b}{HTML}{D7EDFF}

\definecolor{Color4}{HTML}{FCCBEA}
\definecolor{Color4b}{HTML}{FFEAF5}
\definecolor{Color4c}{HTML}{F5DBE6}

\definecolor{Color5}{HTML}{F0F0F0}
\definecolor{Color5b}{HTML}{E0E0E0}

\definecolor{tabindex}{HTML}{888888}

\begin{document}
% \renewcommand\thelinenumber{\color[rgb]{0.2,0.5,0.8}\normalfont\sffamily\scriptsize\arabic{linenumber}\color[rgb]{0,0,0}}
% \renewcommand\makeLineNumber {\hss\thelinenumber\ \hspace{6mm} \rlap{\hskip\textwidth\ \hspace{6.5mm}\thelinenumber}}
% \linenumbers
\pagestyle{headings}
\mainmatter
\def\ECCVSubNumber{6214}  % Insert your submission number here

\title{Webly Supervised Concept Expansion for General Purpose Vision Models}

% INITIAL SUBMISSION 
%\begin{comment}
% \titlerunning{Temp \ECCVSubNumber} 
% \authorrunning{A. Kamath et al. \ECCVSubNumber} 
% \author{Anonymous ECCV submission}
% \institute{Paper ID \ECCVSubNumber}
%\end{comment}
%******************

% CAMERA READY SUBMISSION

\def\@fnsymbol#1{\ensuremath{\ifcase#1\or *\or \dagger\or \ddagger\or
   \mathsection\or \mathparagraph\or \|\or **\or \dagger\dagger
   \or \ddagger\ddagger \else\@ctrerr\fi}}
   
\makeatletter
\newcommand{\printfnsymbol}[1]{%
  \textsuperscript{\@fnsymbol{#1}}
}
\makeatother

\titlerunning{Webly Supervised Concept Expansion for GPVs}
% If the paper title is too long for the running head, you can set
% an abbreviated paper title here
%
\author{Amita Kamath\thanks{Equal contribution}\inst{1} \and
Christopher Clark\printfnsymbol{1}\inst{1} \and
Tanmay Gupta\printfnsymbol{1}\inst{1} \and
Eric Kolve\inst{1} \and
Derek Hoiem\inst{2} \and
Aniruddha Kembhavi\inst{1}}
\authorrunning{A. Kamath et al.}
% First names are abbreviated in the running head.
% If there are more than two authors, 'et al.' is used.
%
\institute{Allen Institute for Artificial Intelligence \and
University of Illinois at Urbana-Champaign
%\email{\{amitak, chrisc, tanmayg, erick, anik\}@allenai.org}\\
%\email{dhoiem@illinois.edu}
}

%******************

\maketitle

\input{sections/0_abstract}

\input{sections/1_intro}

\input{sections/2_relatedwork}
\input{sections/3_dataset}
\input{sections/4_model}

\input{sections/5_dce}
\input{sections/6_experiments}
\input{sections/7_discussion}

% ---- Bibliography ----
%
% BibTeX users should specify bibliography style 'splncs04'.
% References will then be sorted and formatted in the correct style.
%
{\small
\bibliographystyle{splncs04}
\bibliography{derek_bib}
}

\clearpage

\input{sections/11_appendix}

\end{document}

%% file: sections/0_abstract.tex
\begin{abstract}
General Purpose Vision (GPV) systems are models that are designed to solve a wide array of visual tasks without requiring architectural changes. Today, GPVs primarily learn both skills and concepts from large fully supervised datasets. Scaling GPVs to tens of thousands of concepts by acquiring data to learn each concept for every skill quickly becomes prohibitive. This work presents an effective and inexpensive alternative: learn skills from supervised datasets, learn concepts from web image search, and leverage a key characteristic of GPVs: the ability to transfer visual knowledge across skills. We use a dataset of 1M+ images spanning 10k+ visual concepts to demonstrate webly-supervised concept expansion for two existing GPVs
% \footnote{While all existing architectures are limited in the input-output modalities and tasks supported, we refer to them as GPVs to differentiate them from specialized models (both single-purpose and multi-head multitask models) and to acknowledge their ultimate goal of generality.} 
(\gpvone~and \vltf) on 3 benchmarks: 5 \coco-based datasets (80 primary concepts), a newly curated series of 5 datasets based on the OpenImages and VisualGenome repositories ($\sim$500 concepts), and the Web-derived dataset (10k+ concepts). We also propose a new architecture, \gpvtwo\ that supports a variety of tasks --- from vision tasks like classification and localization to vision+language tasks like QA and captioning, to more niche ones like human-object interaction detection. \gpvtwo\ benefits hugely from web data and outperforms \gpvone\ and \vltf\ across these benchmarks. Our data, code, and web demo are available at \url{https://prior.allenai.org/projects/gpv2}.
% \web\ also substantially improves zero-shot performance of \gpvtwo\ on an action recognition benchmark.

%, and shows 0-shot generalization to action and attribute recognition tasks.\todo{Remove 0-shot from abstract ?}

\keywords{General Purpose Vision systems; Webly supervised data}

\end{abstract}

% Decoded abstract for reference:
% General purpose vision (GPV) systems are models that are designed to solve a wide array of visual tasks without requiring architectural changes. Today, GPVs primarily learn both skills and concepts from large fully supervised datasets. Scaling GPVs to tens of thousands of concepts by acquiring data to learn each concept for every skill quickly becomes prohibitive. This work presents an effective and inexpensive alternative: learn skills from supervised datasets, learn concepts from web image search, and leverage a key characteristic of GPVs -- the ability to transfer visual knowledge across skills. We use a dataset of 1M+ images spanning 10k+ visual concepts to demonstrate webly-supervised concept expansion for two existing GPVs (GPV-1 and VL-T5) on 3 benchmarks - 5 COCO based datasets (80 primary concepts), a newly curated series of 5 datasets based on the OpenImages and VisualGenome repositories (~500 concepts) and the Web-derived dataset (10k+ concepts). We also propose a new architecture, GPV-2 that supports a variety of tasks -- from vision tasks like classification and localization to vision+language tasks like QA and captioning to more niche ones like human-object interaction detection. GPV-2 benefits hugely from web data and outperforms GPV-1 and VL-T5 across these benchmarks. 

%% file: sections/1_intro.tex
%\vspace{-0.1in}

\section{Introduction}
\label{sec:intro}

\begin{figure*}[t]
\centering
\includegraphics[width=1.0\textwidth]{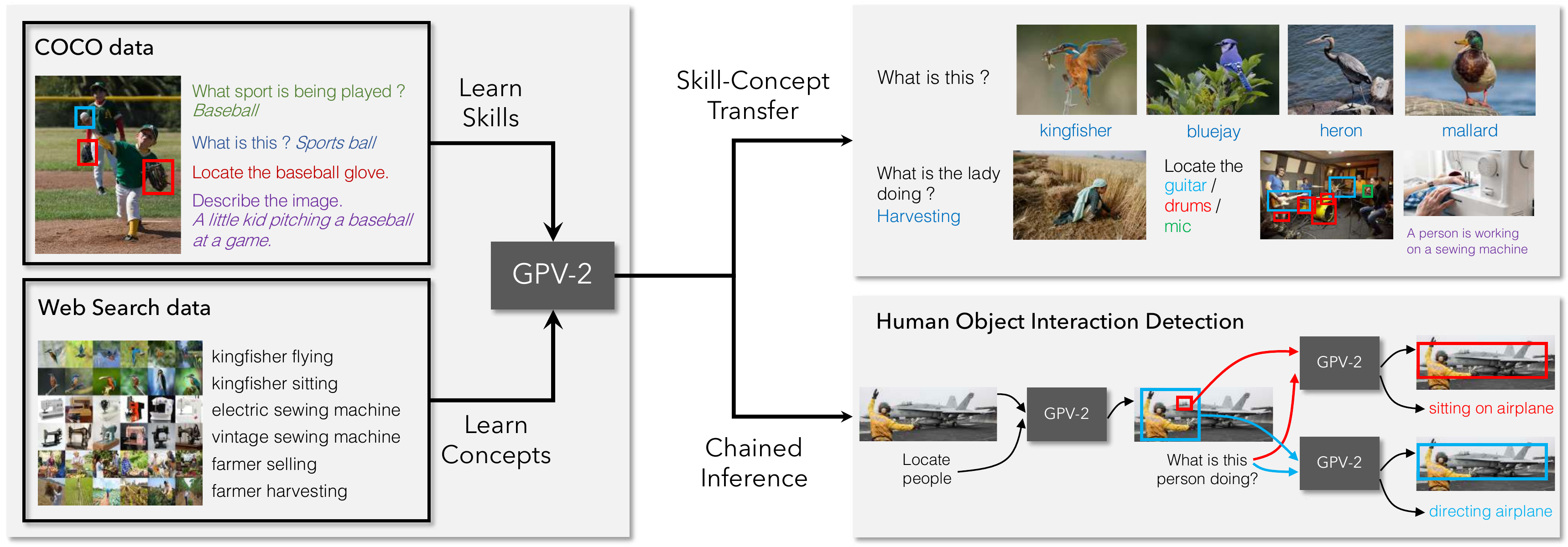}
%\vspace{-0.1in}
\caption{
\textbf{Learning concepts from the web with \gpvtwo.} 
% We propose expanding concept knowledge of general purpose vision systems by learning skills from supervised datasets while learning concepts from web image search data. Existing GPVs effectively transfer webly-supervised concepts across skills such as captioning, classification, and localization. 
We demonstrate webly-supervised concept expansion on two existing GPV architectures (\gpvone\ and \vltf) as well as our proposed \gpvtwo\ architecture. In addition to outperforming previous architectures, \gpvtwo\ expands the inputs to contain bounding boxes which enables support for niche tasks like Human-Object Interaction detection with multi-step inference without any architectural modifications.\label{fig:teaser}}
%\vspace{-0.3in} %%
\end{figure*}

General Purpose Vision systems (GPVs) \cite{gpv1} are designed to support a wide range of tasks without requiring architectural changes. A task is the application of skills (e.g. localization, captioning) to concepts (e.g. monkey, brown, climbing) in order to map from the input (image, text) to a target output (text, boxes).  Given the virtually unlimited number of fine-grained and topical concepts, it is not feasible to provide a GPV with annotations for all skills on all concepts, as even large pre-collected datasets  cannot anticipate every need. In this work, we ask: \emph{Can a GPV leverage web image search and skill-concept transfer to massively and inexpensively expand its concept vocabulary across a variety of tasks?} To answer this question, we present a large-scale webly supervised dataset for learning 10k+ concepts, a new benchmark for broader concept evaluation ($\sim$500) across 5 diverse tasks, and a new GPV architecture that improves cross-task concept transfer and outperforms existing GPVs across multiple benchmarks.

%While specialized models are designed to learn from datasets collected for their specific task, GPVs aim to unlock the potential of learning skills and concepts from a wide range of task-datasets and then generalize those learned skills and concepts to novel skill-concept combinations, such as being able to localize and answer questions about ukeleles after learning to classify images into ``ukelele''. Scaling GPVs to tens of thousands of concepts, however, remains a challenge, due to the need for expensive supervised datasets that only expose the model to a limited number of concepts for each skill (e.g. 80 \coco~\cite{Lin14Microsoft} categories). In this work, we ask - \emph{Can we leverage web image search and the skill-concept transfer ability of a GPV to massively and inexpensively expand its concept vocabulary across a variety of tasks ?} To answer this question, we present a large scale webly supervised dataset for learning 10k+ concepts, a new benchmark for broader concept evaluation ($\sim$500) across 5 diverse tasks, and a new GPV architecture that improves cross-task concept transfer and outperforms present day GPVs across multiple benchmarks.

Image search engines provide remarkably good results for millions of queries by leveraging text on the accompanying web pages, visual features from images, and click data from millions of users querying and selecting relevant results each day. They often provide high-quality, decluttered, object- and action-centric images, which can be used to learn powerful visual representations for concepts. Importantly, searches scale easily and inexpensively to thousands of queries. Given the large cost of producing high-quality supervised datasets, scaling today's manually annotated datasets to support 10,000+ concepts is infeasible for many tasks. In contrast, using Bing search to create \web, a dataset with 1M+ images spanning 10k nouns, 300 verbs, and 150 adjectives with thousands of noun-verb and noun-adj combinations, cost us just over \$150. Moreover, while existing data sources such as ImageNet-22k and YFCC100M are valuable resources, they are static snapshots of a diverse and ever-changing world. For example, these static datasets may not represent specialized categories of interest to a downstream application such as \emph{boysenberry} and will definitely not contain latest concepts such as \emph{Pixel 6} or \emph{COVID-19 home test}. On the other hand, modern web image search engines are designed to serve imagery on-demand and are uniquely positioned to act as a source of training data for novel and latest concepts. While search engine data provides strong supervision for classification, we demonstrate that current GPVs, \gpvone~\cite{gpv1} and \vltf~\cite{vlt5}, are able to learn concepts from web data and improve on other skills as well, such as image captioning. Importantly, we show that even models that already utilize large-scale pretraining corpora such as Conceptual Captions continue to benefit from using search engine data and can be easily extended to support new concepts relevant in the present day that have little or no coverage in large static corpora.

We also propose \gpvtwo, a powerful GPV %with support for a broader set of modalities (and hence tasks) beyond those supported by present day GPVs. 
that can accept as input an image, a task description, and a bounding box (allowing the user to point at an object or region of interest), and output boxes and text for any bounding box or for the entire image. These diverse input and output modalities enable \gpvtwo\ to support a large spectrum of skills ranging from vision skills like classification and localization, vision-language skills like VQA and captioning, to niche ones like classification in context and human-object interaction detection. An important design principle of \gpvtwo\ is  Language-Based Localization, whereby {\em all} tasks are based on scoring/ranking/generation using the same text decoder applied to one or more image regions. This ensures that all tasks share the same weights and representations, ranging from the input encoders all the way to the output decoders --- resulting in more effective skill-concept transfer for learning from diverse tasks' datasets. We also propose a re-calibration mechanism to downweight scores of labels that are disproportionally represented in training, and demonstrate its effectiveness on out-of-domain test datasets for multiple tasks.

Benchmarking the diverse capabilities of large-vocabulary general purpose models is challenging. Most current datasets in computer vision are designed for single tasks. The recently proposed \cocosce~\cite{gpv1} benchmark is designed to test the skill-concept transfer ability and overall skill competency across five vision skills. However, it is limited to evaluate these competencies on 80 primary \coco\ concepts. In this work, we present a new benchmark named \opensce\ for broader concept evaluation for the same five skills but now expanding to 492 \openim\ concepts. \opensce\ is an evaluation-only benchmark sourced from \openim~\cite{Kuznetsova18OpenImages}, \vg~\cite{Krishna17Visual} and NoCAPS~\cite{Agrawal2019nocapsNO} with new VQA annotations and has been sampled in a way that prevents over-representation of any single category while maximizing representation of infrequent categories.

We evaluate present day GPVs and \gpvtwo\ on three benchmarks: (i) the \cocosce\ and \coco\ benchmarks~\cite{gpv1}, (ii) the newly presented \opensce\ benchmark; and (iii) the \web~dataset consisting of \emph{manually verified} images from Bing Image Search paired with questions and answers that covers 10,000+ concepts. Our analysis shows that all three GPVs benefit from web data. Furthermore, \gpvtwo\ outperforms both \gpvone\ and \vltf\ across these benchmarks and shows significantly large gains when using web data, particularly for captioning and classification. 
% \gpvtwo~also performs well at downstream tasks like action and visual attribute recognition. 
We also demonstrate how \gpvtwo\ can be chained to perform niche tasks like human-object interaction detection, without any task-specific architecture modifications. Finally, we show how web data can be efficiently used to expand \gpvtwo's concept vocabulary to include new visual concepts that are relevant in today's world such as \emph{COVID-19 vaccination cards} and \emph{N95 masks}, concepts that are infrequent or non-existent in static corpora.

In summary, our main contributions include: (a) \web, a new web data source to learn over 10k visual concepts with an accompanying human-verified VQA benchmark; (b) demonstration that GPVs can learn concepts from \web\ and transfer this knowledge to other tasks;
(c) \opensce, a benchmark spanning 5 tasks and approximately 500 concepts to evaluate GPVs; and (d) \gpvtwo, an architecture that supports box and text modalities in both input and output, improves skill-concept transfer and outperforms existing GPVs.
% by sharing the same encoders and decoder for all tasks and using classifier re-calibration and outperforms existing GPVs,
% and achieves reasonable zero-shot generalization to visual attribute and verb recognition tasks. 
Our code and benchmarks are available at \url{https://prior.allenai.org/projects/gpv2}, along with a new tool to easily create a web dataset from a list of queries.

%% file: sections/2_relatedwork.tex
\section{Related Work}
\label{sec:related_work}

\oursubsection{General purpose models}
Computer vision models have progressively become more general. Specialization first gave way to multitask models which aimed at solving multiple, albeit predefined, tasks with one architecture. A common approach for building such models~\cite{Lu202012in1MV,He17Mask} is to use task-specialized heads with a shared backbone. However, adding a new head for each new task makes scaling to a large number of tasks and reuse of previously learned skills challenging. 
An alternative approach is to build a \textit{general-purpose} architecture without task-specific components. This approach has become common in natural language processing via text-to-text generative models~\cite{Raffel2020t5,Brown2020gpt3,McCann2018decanlp}, and recent work in computer vision has striven towards this kind of generality~\cite{Dosovitskiy2021ViT,Carion2020DETR,Kim2021ViLTVT,Liu2021SwinTH}. 
% On the other hand, in the NLP domain, transformer based models such as T5~\cite{Raffel2020t5} and GPT-3~\cite{Brown2020gpt3} made it possible to use the same architecture for almost any NLP task without any modification. 
% With the rising popularity of transformer models in vision~\cite{Dosovitskiy2021ViT,Carion2020DETR,Kim2021ViLTVT,Liu2021SwinTH}, several recent works are striving towards this generality in computer vision as well. 

Examples of general-purpose computer vision models include VL-T5~\cite{vlt5}, which adapts T5~\cite{Raffel2020t5} to jointly train on vision+language (V+L) tasks while using a single text-generation head to produce outputs for all tasks, and GPV-1~\cite{gpv1}, which combines a similar text-generation head with the ability to return bounding-boxes and relevance scores as output.
% While collapsing multiple output heads in vision-language models, the architecture mainly focuses on tasks with a text output and formulates the referring expression task as referring to a precomputed object region in the image via text. This approach has not been tested for tasks in vision such as object localization. 
In this work, we work with both \gpvone\ and \vltf\ and extend their concept vocabulary with web data. Our proposed model, \gpvtwo\ follows \vltf\ in its use of the T5 backbone, builds upon the vision capabilities of \gpvone, and further extends the range of tasks that can be performed by allowing a bounding-box input and introducing the ability to generate per-image-region text output. 
Perceiver~\cite{perceiver} and PerceiverIO~\cite{perceiverIO} aim to generalize the architecture beyond images and text to other modalities such as audio, video, and point cloud. However, both architectures remain to be tested for multitask learning and for learning V+L tasks such as VQA and captioning.
Many other V+L models~\cite{Xu2021E2EVLP,Tan2019LXMERTLC,Chen2019UNITERLU,Li2020OscarOA,Lu20Multitask} can be fine-tuned on a variety of downstream tasks, but they typically use task-specific heads, while the focus of our work is on general purpose models in a multi-task setting.

% Other multi-task vision models include E2E-VLP~\cite{Xu2021E2EVLP}, which has an end-to-end trainable architecture that is pretrained with image-text and bounding box annotations. The pretrained model is then finetuned on each task separately instead of sharing weights across tasks. 

% Need to cite and talk about:
% \begin{enumerate}[noitemsep]
%     \item Specialized architectures
%     \item Models with common backbones but specialized heads and trained separately
%     \item Models with common backbones that are trained jointly
%     \item General purpose models / architectures
% \end{enumerate}

% General purpose architectures:
% \begin{itemize}[noitemsep]
%     \item \gpvone~\cite{gpv1} -- Our closest competitor
%     \item VL-T5~\cite{vlt5} -- A general purpose model focused on V\&L tasks
%     \item Perceiver~\cite{perceiver} - A unified encoder mechanism based on transformers that works for images, text and audio
%     \item PerceiverIO~\cite{perceiverIO} - Extends Perceiver to both inputs and outputs
% \end{itemize}

\oursubsection{Web supervision}
Image search engines provide highly relevant results, using a combination of text, image and user features.
%use not just image features, but also rely heavily on text on web pages as well as the power of millions of users clicking on relevant results to surface highly relevant image results, particularly for common user queries. 
Researchers have used search data as a form of supervision to build computer vision models. Early works used noisy retrieved results with probabilistic Latent Semantic Analysis~\cite{Fergus2005LearningOC} and multiple instance learning~\cite{Vijayanarasimhan2008KeywordsTV} to build recognition systems. As web results improved, works used this data to build object detectors ~\cite{Divvala14Learning,Chen15Webly,Li2013HarvestingMV,Shen2020NoiseAwareFW,Luo2020WeblysupervisedLF,Wu2020ExploringBA}, attribute detectors~\cite{Golge2014ConceptMapMN}, image taggers~\cite{Wang2008AnnotatingIB}, large vocabulary categorization models~\cite{Yang2020WeblySI,Niu2018LearningFN,Guo2018CurriculumNetWS} and fine-grained recognition models~\cite{Krause2016TheUE,Niu2018WeblySL}, segmentation models~\cite{Shen2018BootstrappingTP,Jin2017WeblySS,Sun2020MiningCS}, online dataset builders~\cite{Li2007OPTIMOLAO}, visual reasoning systems~\cite{Zheng2020WeblySK} and visual knowledge bases with learnt relationships between objects~\cite{Chen13NEIL}. More recently, massive scale web data in the form of retrieved search results and the accompanying text was employed to build the powerful CLIP family of models~\cite{radford2021learning_CLIP_arxiv} that provide powerful visual representations for downstream tasks. While these works have shown that web data can be used to build single task models, we show that one can build GPVs with web data and importantly transfer this knowledge across skills.

\oursubsection{Concept transfer across skills}
There has been considerable interest in transferring concept knowledge from classification to object detection, as classification labels are far cheaper to obtain than detection labels. Hoffman \etal~\cite{Hoffman2014LSDALS} cast this problem as a domain adaptation problem, adapting classifiers to detectors. Redmon \etal~\cite{Redmon2017YOLO9000BF} build a 9,000 class detector using Imagenet22k classification data~\cite{Dong09ImageNet} by jointly training for the two tasks. Uijlings \etal~\cite{Uijlings2018RevisitingKT} use Multiple Instance Learning to pseudo-label data and train a large vocabulary detector. Recent works build open vocabulary detectors~\cite{Zareian2021OpenVocabularyOD,gu2021openvocabulary,Jia2021ScalingUV} by leveraging image caption pairs (or models like CLIP~\cite{clip} which are built from the same), obtained in large quantities on the web. Even though image-captions are noisy, the resulting detectors improve as the data is scaled up.

The V+L field has leveraged object detectors as feature inputs~\cite{Anderson2018BottomUpAT,Zhang2021VinVLMV,Agrawal2019nocapsNO}, which can be considered as transferring concepts from detection to downstream tasks. Another effective approach is pre-training using image-captions~\cite{Lu2019Vilbert,Li2019VisualBERTAS,Li2020OscarOA} like Conceptual Captions~\cite{Sharma18Conceptual}. CLIP~\cite{clip} is a family of powerful models that are pre-trained on a massive 400M image caption paired dataset. The resulting encoders are very effective at V+L tasks~\cite{Shen2021HowMC}. These methods effectively transfer visual knowledge from caption data to tasks like VQA. Recently Whitehead \etal~\cite{Whitehead2021SeparatingSA} disentangle the encoding of concepts and skills and build a model that can generalize to new skill-concept compositions and new concepts for VQA.

The focus of our work is to build a GPV that can transfer concepts across various skills, particularly from web data to vision and vision-and-language skills, and also provide a new test-only evaluation benchmark for the same.

%The focus of our work is to create an inexpensive, highly scalable web dataset that GPVs can leverage to expand their knowledge of a large number of concepts across various skills.  

%% file: sections/3_dataset.tex
\section{The WEB10K dataset}
\label{sec:dataset}
% \todo{Do mention licenses here? We state we will do in paper question 6}
% \input{tables/web_stats}

% \begin{figure*}[ht]
% \begin{center}
% \includegraphics[width=1.0\textwidth]{figs/web_figure.pdf}
% \end{center}
% \caption{\textbf{\web}. }\label{fig:web}
% \end{figure*}

% \begin{figure}[h]
% \begin{center}
% \includegraphics[width=0.9\linewidth]{figs/web_figure_vertical.pdf}
% \end{center}
% \caption{\textbf{\web}. }\label{fig:web}
% \end{figure}

\begin{figure*}[t]
\begin{center}
\includegraphics[width=1.0\textwidth]{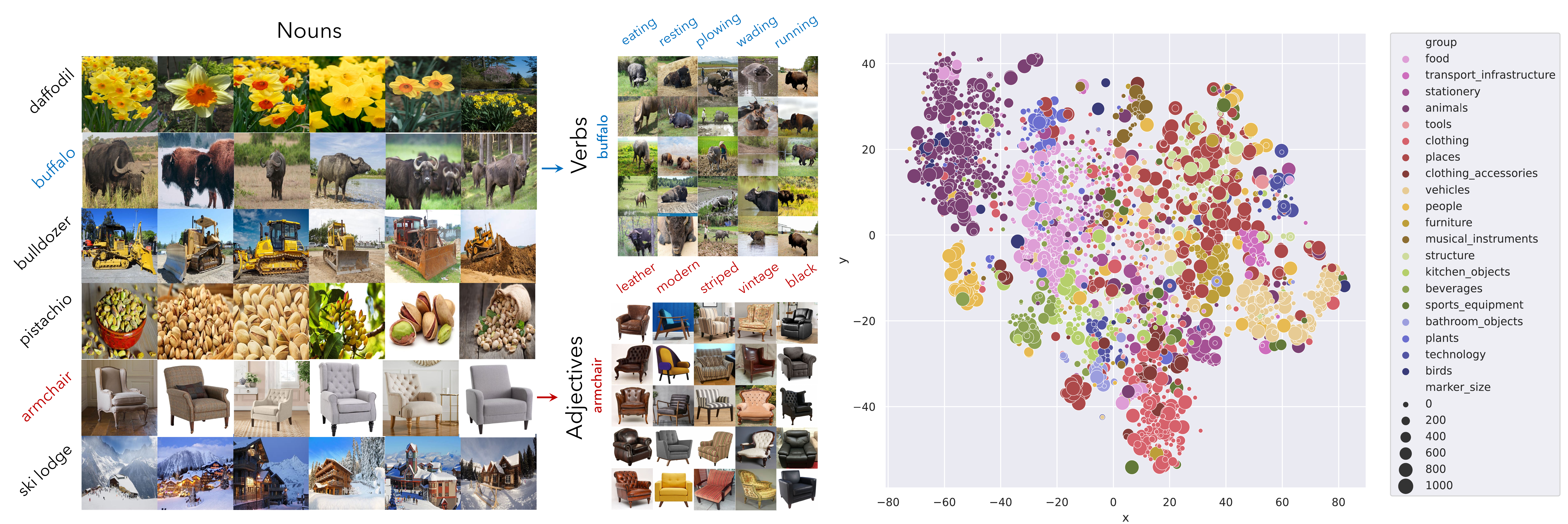}
\end{center}
%\vspace{-0.15in}
\caption{\textbf{Concept diversity in WEB10K.} \textbf{Left:} Besides 10k nouns, \web\ provides dense coverage of feasible adj-noun and verb-noun combinations to enable learning of fine-grained differences in object appearance due to attributes. 
% Conversely, most verbs and adjectives appear as attributes of multiple objects to help models disentangle objects from attributes. 
\textbf{Right:} TSNE~\cite{Maaten2008tsne} plot of Phrase-BERT~\cite{Wang2021PhraseBERT} embeddings of \web~nouns with bubble size indicating frequency (capped at 1000) in \cc, a common large-scale pretraining dataset. \web\ nouns cover a wide range of concept groups identified using WordNet and include many concepts which are infrequent/absent in \cc.}
%\vspace{-0.2in} %%
\label{fig:web}
\end{figure*}

Search engines can be leveraged to collect datasets with highly desirable characteristics: (1) \textbf{Diversity} --- Search engines benefit from a large volume of user click data to produce high-quality results for a large vocabulary of concepts including tail concepts not frequently mentioned in annotated computer vision datasets (e.g. \emph{hyacinth}); (2) \textbf{Freshness} --- Search engines are designed to serve the freshest content on the internet, and often produce very good results for the latest queries (that may not have existed or been popular before; e.g. \emph{COVID-19 vaccination card}, \emph{2022 winter olympics mascot}) which have few/no occurences in standard vision datasets that tend to be static; and (3) \textbf{Concept focus} --- The image distribution of search engine results tends to be similar to image classification data with the image centered on the queried object with few distractions, making them ideal for learning visual concept representations.

We present \web, a dataset sourced from web image search data with over 10K concepts. \web\ contains queries with nouns, adjectives and verbs.
% This dataset can thus be used to introduce a large number of clean concepts to the model, for transfer to other tasks.

% This section discusses data collection, the creation of a QA dataset based on the query-image pairs, and manual verification of the same.

\noindent \textbf{Nouns.} We consider single and multi-word nouns. Single-word nouns are sourced from a language corpus with a list of 40,000 concrete words~\cite{Brysbaert2013Concreteness}, each with a concreteness score (defined as the degree to which a word refers to a perceptible entity).
% and word type (primarily POS tags but also subcategories like Name and Number). 
% Brysbaert \etal~\cite{Brysbaert2013Concreteness} define Concreteness as the degree to which a word refers to a perceptible entity. 
From this list, we select nouns with a concreteness score $> 4.0/5$ and any verb or adjective with an alternate word sense as a noun (e.g. “comb”) with a score $> 4.5/5$. These thresholds avoid more abstract or non-visual words such as ``humor''.
%Words corresponding to nouns are selected if this score $> 4.0/5$, and for verbs and adjectives with an alternate word sense as a noun (e.g. “comb”), we set a bar of $> 4.5/5$. These high bars help us avoid nouns that are difficult to visualize, e.g. ``humor''. 
Multi-word nouns are sourced from \ccfull\ (\cc)~\cite{Sharma18Conceptual}. We identify candidates using POS tagging and select the most frequent 2,000, and an additional 282 where the second word of the multi-word noun is present in the concreteness dataset (e.g. ``street artist'', where ``artist'' is in concrete nouns). In total, we select 10,211 nouns. Sourcing nouns from a language corpus enables coverage of concepts not commonly covered in vision datasets: over 4,000 nouns in \web\ are not present in \cc, e.g. ``wind tunnel''. 

\noindent \textbf{Verbs.} We source verbs from a combination of vision datasets with large verb vocabularies including imSitu~\cite{Yatskar2016SituationRV}, HICO~\cite{ChaoHOIBench} and VRD~\cite{Liang2018VisualRD}. We remove verbs that are either polysemous (have multiple meanings e.g. “person holding breath” vs. ``person holding cup") or aren't associated with an animate agent (e.g. “snowing”). %, or are ambiguous without specifying the whole subject-verb-object triple (e.g. “man holding breath” vs. ``man holding cup"). 
This results in 298 unambiguous and visually recognizable verbs. These verbs improve model performance on action recognition (Supplementary Sec. 8).

%We filtered out the ones with multiple meanings (polysemy), the ones that aren't associated with an animate agent (e.g. “snowing”), and the ones that are ambiguous without specifying the whole subject-verb-object triple (e.g. “man holding breath” vs ``man holding cup"). %This results in 298 verbs, most of which are clear to identify visually and tend to be unambiguous.

\noindent \textbf{Adjectives.} We source adjectives from several datasets that have a large number of adjectives~\cite{Sharma18Conceptual,Krishna17Visual,Farhadi2009DescribingOB,Lampert2009LearningTD,Kumar2009AttributeAS,Patterson2012SUNAD,Parikh2011RelativeA,Chen2012DescribingCB,Wang2013WeaklySL}. 
% Care was taken to filter out ones that were subjective, non-visual or relative. This results in 148 adjectives which were then grouped into 16 adjective types (e.g. “color”).
We manually filter out ones that are subjective (“beautiful”), non-visual (“loud”), or relative (“big”). This results in 144 adjectives which we group into 16 adjective types (e.g. “color”, “texture”).

We select noun-adj pairs and noun-verb pairs which appear at least thrice in \cc: this removes nonsensical pairs, e.g. ``cloudy dog''. 
The total number of queries in \web\ is 38,072 with roughly 10k nouns, 18k noun-adj and 9k noun-verb combinations. We feed each query into the Bing Search API and retrieve a total of 950,443 image URLs (approx. 25 per query). 
% \todo{I'm not sure where to fit this in: We retrieved the Bing thumbnail of each image, as the size was sufficient and the thumbnails tended to have fewer watermarks.} 
\textbf{Importantly, this cost us \$154}, so it is inexpensive to scale further, and such data acquisition is affordable for many other research organizations. %While certainly not an insignificant amount, this manageable cost allows us and other researchers to inexpensively scale this data further, and makes massive and relatively clean data acquisition accessible to many other research organizations.
See Tab.~\ref{tab:dce_stats} for detailed statistics.

\oursubsection{Conversion into QA data} We convert each query-image pair into multiple templated QA pairs where the answer is the noun, adjective or verb from the query. For example ``What is the $[$noun$]$ doing?'' and ``What $[$adj type$]$ is this object?''; see Supplementary Sec. 3 for all question templates. The QA format has two advantages: (1) it removes ambiguity from the task (e.g., ``What color is this" tells the model not to return a potentially accurate non-color attribute); and (2) it bridges the domain gap to other tasks posed as questions. 

% By training the decoder to generate tail category words for this task, it becomes less biased against generating them at inference time. 
\oursubsection{Data Splits} We split image-query pairs into train (874k), val (38k) and test (38k). We sample 5k and 10k pairs from the val and test sets and ask 3 crowdworkers to verify that the query is present in the image. We only retain unanimously verified examples (71\%) resulting in: Val -- 4k images (9k QAs), Test -- 8k images (19k QAs). The Train set has about 3M QAs with no manual verification.

%\begin{itemize}[noitemsep]
%    \item Sourcing noun, verb and adjective lists
%    \item Filtering lists based on freq, intersection with other lists, concreteness, etc
%    \item Bing API details including cost
%    \item Statistics of the dataset
%    \item Question types
%    \item Train / Val / Test sets
%    \item Manual verification / clean-up of val and test sets
%    \item Dataset figure
%\end{itemize}

\input{tables/dce_stats}

%% file: tables/dce_stats.tex
\begin{table}[t]
\setlength{\tabcolsep}{5pt}
\scriptsize
\centering
\caption{
\label{tab:dce_stats}
\textbf{Left:} \web\ statistics (Sec.~\ref{sec:dataset}). There are approximately 25 images per concept.
\textbf{Right:} \opensce\ val and test statistics (Sec.~\ref{sec:dce}).  
% Since nocaps~\cite{Agrawal2019nocapsNO} annotations are hidden behind an evaluation server, we are unable to provide category counts for captioning.
}
\resizebox{\textwidth}{!}{
\begin{tabular}{cl}
        \toprule
        Type & Count \\
        \midrule
      \multirow{6}{*}{Concepts } &  Nouns: 10211 \\
      & Adjectives: 144 \\ 
      & Verbs: 298 \\
      & Noun-adjective pairs: 18616 \\
      & Noun-verb pairs: 9243 \\
      & \textbf{Total: 38072} (Nouns + Pairs) \\
      \midrule
      \multirow{4}{*}{Images} &  Noun images: 255073 \\
       & Noun-adjective images: 465146 \\
       & Noun-verb images: 230224\\
       & \textbf{Total: 950443} \\
       \midrule
      \multirow{5}{*}{QAs} & Templates: 26 \\
& Noun QAs: 1900886 \\
& Adjective QAs: 930292 \\
& Verb QAs: 460448 \\
& \textbf{Total: 3291626} \\
        \bottomrule
    \end{tabular}
\quad
\quad
\begin{tabular}{llccc}
\toprule
Subset & Skill & Samples & Images & Categories\\ %\hline
\midrule
\multirow{4}{*}{Val}  & VQA            & 5169  & 2131  & 295 \\
                      & Localization   & 8756  & 7588  & 463 \\
                      & Classification & 9485  & 6770  & 464 \\
                      & Cls-in-context & 9485  & 6770  & 464 \\
                      & Captioning     & 4500  & 4500  & -   \\
\midrule
\multirow{4}{*}{Test} & VQA            & 5281  & 2160  & 307 \\
                      & Localization   & 10586 & 9986  & 476 \\
                      & Classification & 10888 & 9161  & 476 \\
                      & Cls-in-context & 10888 & 9161  & 476 \\
                      & Captioning     & 10600 & 10600 & -  \\
\bottomrule
\\
\multicolumn{5}{l}{Note: Since nocaps~\cite{Agrawal2019nocapsNO} annotations are hidden behind an}\\
\multicolumn{5}{l}{evaluation server, we are unable to provide category}\\
\multicolumn{5}{l}{counts for captioning.}\\
\end{tabular}}
\end{table}

%% file: sections/4_model.tex
\section{\gpvtwo}
\label{sec:model}

In this section we present \gpvtwo, a model combining an object detector with the T5 pre-trained language model. \gpvtwo\ supports additional input and output modalities (and thus tasks) beyond present day GPVs (\gpvone\ and \vltf). It uses the stronger VinVL ~\cite{Zhang2021VinVLMV} object detector, uses a shared language decoder (for all tasks including localization) and employs a classification re-calibration approach, which together improve generalization to unseen concepts at test time.

\begin{figure*}[t]
\centering
\includegraphics[width=1.0\textwidth]{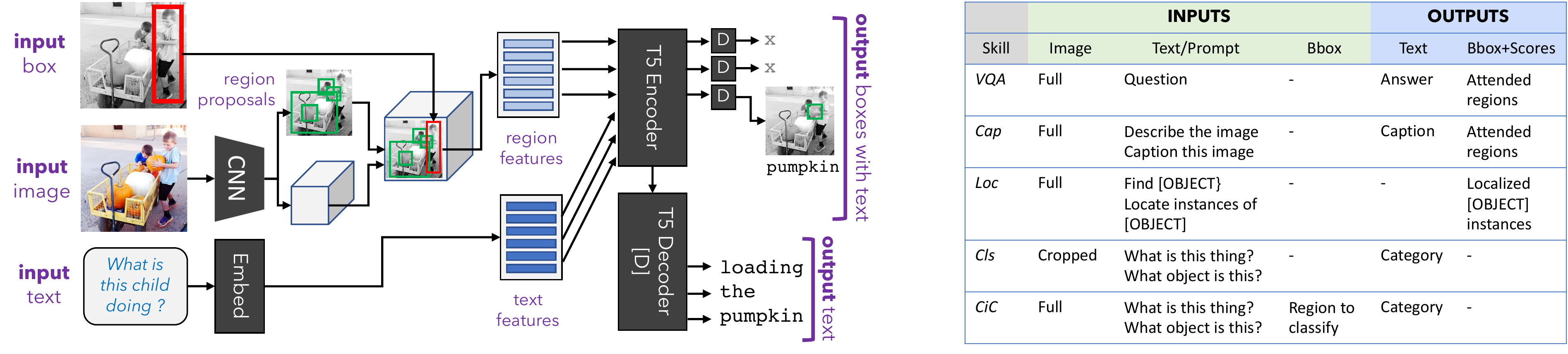}
%\vspace{-0.1in}
\caption{
\textbf{Left}: \gpvtwo\ architecture. \textbf{Right}: I/O for 5 skills in \coco\ and \opensce.
}
%\vspace{-0.2in} %%
\label{fig:model_io}
\end{figure*}

% \begin{figure}[h]
% \centering
% \includegraphics[width=0.6\linewidth]{figs/gpv2_model.pdf}
% \caption{
% The \gpvtwo\ model architecture.
% }
% \label{fig:model}
% \end{figure}

\oursubsection{Model design}
% \subsection{Model Design}
\gpvtwo\ takes an image, text, and a bounding box as input. As output, it can produce text for an individual bounding box (the input box, or boxes produced by the visual model) and for the entire image (see Fig.~\ref{fig:model_io}). 

First, the input text is tokenized and embedded using T5-Base to get a sequence of text feature vectors. Then an object detection model is used to identify regions in the image and extract bounding boxes and features for those regions (we do not use the class labels identified by the detector) via RoI pooling. We additionally use the object detector to extract features for the input bounding box, and a learned embedding is added to those features to distinguish them from the other visual features. These sets of visual features are then converted to embeddings of the same dimensionality as the text embedding using a linear layer. We primarily use the VinVL~\cite{Zhang2021VinVLMV} object detector for our experiments. 
% However the \gpvtwo\ architecture allows us to easily swap in other detectors (see  Sec.~\ref{sec:exp}).
However the \gpvtwo\ architecture allows us to easily swap in other detectors, and we use features from DETR~\cite{Carion2020DETR} for some of our experiments in Sec.~\ref{sec:exp}.

The resulting visual and text vectors are concatenated as a sequence and used as an input to the T5-Encoder to build joint contextualized embeddings. To generate text for the entire image we use the T5-Decoder with this contextualized embedding sequence as input, and to generate text for individual boxes we run the same T5-Decoder while using the contextualized embedding that corresponds to just that box as input. The usage of a common decoder for image-based outputs and region-based outputs enables transfer of learned concepts between skills that require processing the entire image and skills that rely primarily on the representation of a single region. 

\oursubsection{Using \gpvtwo}
% \subsection{Using \gpvtwo}
% \label{sec:gpv2-outputs}
\gpvtwo's design gives us flexibility to handle a variety of vision and vision+language tasks without needing task-specific heads. 
For tasks that do not have text input, we follow \cite{gpv1} by building appropriate text prompts for that task (e.g., \prompt{What is this object?} for classification) and selecting one at random to use as the input text. For tasks that do not have an input bounding box, we use a box around the entire image. 

Decoded text from the image is used to answer questions and generate captions.  For classification or limited-choice responses, answers are scored based on log-probability of generating each option, and the highest scoring answer is chosen. To localize objects, we propose Language-Based Localization (LBL) where a box is scored by first computing the log-probabilities of generating an object class or ``other'' from that box, and then applying a linear classifier to those scores to yield a scalar relevance score.  For example, \prompt{Localize dog} is performed by computing the log-probability of ``dog'' and ``other'' for each region.

Importantly, the same text decoder is used to generate image and region text, thus {\em classification, question answering, captioning, localization, and all other tasks use the same encoders, decoder, and weights}. Our experiments show that this facilitates skill-concept transfer.

Even complex tasks like human-object interaction (HOI) can be performed by chaining inference steps (Fig.~\ref{fig:teaser}).  HOI~\cite{ChaoHOIBench,ChaoHOIDet} requires localizing a person, an object and categorizing their interaction.  \gpvtwo\ performs this by first returning detections for \prompt{Locate person}, then providing each person box as input with the prompt \prompt{What is this person doing?}  The log-probs of generating object-interaction phrases, such as ``directing the airplane'' for other boxes are used to identify the most likely interaction.

\oursubsection{Classification re-calibration}
% \subsection{Classification re-calibration}
% \label{sect:classification-calibration}
We observe that a common issue in classification is that the model becomes biased towards classes that are common in the training data. For example, we find that if the model is trained to classify \coco\ objects it will almost always guess the names of \coco\ objects in response to the prompt \prompt{What is this object?}, even if no such objects exist in the image. This can be viewed as a language bias, as has been well-studied in VQA~\cite{goyal2017cvpr_balanced_vqa_v2,ramakrishnan2018overcoming}. To solve this issue we re-calibrate the models output prediction by reducing the log-probability of classes that were seen in the training data when doing answer re-ranking. The down-weighting amount is selected on the validation data. See Supplementary Sec. 2 for an analysis and example.

\oursubsection{Pre-training}
% \subsection{Pre-training}
Recent works have shown that pre-training V+L models on large amounts of data results in large improvements~\cite{vlt5,Li2020OscarOA,Zhang2021VinVLMV}. We do not have the resources to fully-replicate these setups, but as a partial measure we pre-train \gpvtwo\ for 8 epochs on the \cc 3M dataset~\cite{Sharma18Conceptual}, which shows significant gains on our benchmarks. Since \gpvtwo\ is generative, we pre-train it by simply learning to generate the target caption rather than using span masking or other more complex objectives~\cite{Li2020OscarOA,Tan2019LXMERTLC}. While we use much less data than some V+L works, pre-training on \cc 3M allows us to verify that \gpvtwo\ still benefits from web data even when exposed to a broad range of concepts during pre-training.

% \todo{We also use a much simpler/more natural pre-training objective. Explain the implication/result ?} \todo{We also do not pre-train on datasets. We use a single large resource for pre-training. Why / Implication ?}

%% file: sections/5_dce.tex
\section{\opensce\ Benchmark}\label{sec:dce}
The \coco\ benchmark focuses on 80 object categories and is insufficient for evaluating skills on a wide range of concepts. We introduce the \textbf{\underline{D}}iverse \textbf{\underline{C}}oncept \textbf{\underline{E}}valuation (\opensce) benchmark to evaluate GPV models on a large subset of the 600 \openim\ categories across 5 skills: classification (Cls), classification-in-context (CiC), captioning (Cap), localization (Loc), and visual question answering (VQA). See Fig.~\ref{fig:model_io} for the inputs and outputs for each task. We introduce CiC as a natural and unambiguous object classification task (similar to pointing at an object and asking what it is), providing a direct complement to localization.
%\todo{Motivate why we are introducing the new task of CiC}. 
We source Cls, CiC and Loc samples from \openim, VQA samples from \vg~(VG), and use the nocaps~\cite{Agrawal2019nocapsNO} benchmark for Cap evaluation. To curate the \opensce\ benchmark, we first select a set of mutually exclusive categories from \openim\ and draw samples for each of those categories according to a sampling strategy that prevents over-representation of any category while maximizing representation of tail categories. \opensce\ is an evaluation-only benchmark and is not accompanied by a distributionally similar training set. %\todo{We should add that this is not a train benchmark, merely a test benchmark}

\oursubsection{Category selection} \openim\ provides a total of 600 hierarchical object categories. After removing some categories due to label noise, we use the remaining 492 leaf nodes in the hierarchy as our mutually exclusive set of categories. %\todo{What about category cleanup with regards to parent-child nodes / deduping / etc.? Tanmay: taking leaf nodes in the hierarchy takes care of this, and removal of high label noise categories is by manual inspection so not much else to say here}

\oursubsection{Sampling strategy} For Cls, CiC and Loc, we randomly sample up to 25 samples from each of the selected categories. A sample for Cls/CiC is defined as any bounding box annotated with a category. For Loc, a sample is all bounding boxes in an image annotated with a category (we discard ``group" annotations). For VQA, we first discard annotations exceeding 2 word answers after removing articles and tag each QA pair in VG with any of the selected categories mentioned in either the question or answer. Then, for each category, we sample up to 50 data points. Since each sample in VQA may consist of multiple categories, this strategy does result in more than 50 samples for some categories, but in practice it achieves the goal of preventing common categories from dominating the evaluation. Finally, some of the 492 categories do not have annotations in the source datasets. The final sample, image, and category counts for each skill are in Tab.~\ref{tab:dce_stats} and category frequencies are shown in Supplementary Sec. 4.

%Iterating over the selected category list, we randomly sample up to 50 samples for each category. Unlike Cls/CiC and Loc, each sample in VQA may consist of multiple categories. If $k$ samples have already been sampled for the $i^{th}$ category in the selected category list due to co-occurrence with previous $i-1$ categories, we only sample $\max(0,50-k)$ samples for the $i^{th}$ category. While this greedy sampling strategy does result in exceeding the 50 samples limit for some categories due to co-occurrence, in practice it does achieve the goal of preventing common categories from dominating the evaluation. Finally, some of the 492 categories do not have annotations in the source datasets. The final sample, image, and category counts for each skill are in Tab.~\ref{tab:dce_stats} and category frequency histogram is shown in Supp.

\oursubsection{Additional VQA annotations} VQA annotations from VG only consist of one answer per question. For each selected VQA sample, we source 9 additional answers from Amazon Mechanical Turk as in standard \coco-based VQA benchmarks~\cite{goyal2017cvpr_balanced_vqa_v2,Antol15Visual}. Samples where $\geq$3 workers agreed on an answer were retained. %\todo{What about filtering questions that did not have consensus? Did we do any other filtering post Turk ?}

%% file: sections/6_experiments.tex
\section{Experiments}
\label{sec:exp}

\input{tables/tab_concept_scaling}

We train models jointly on all tasks that are supported by each GPV using \coco-based datasets.
% \gpvtwo\ supports all 5 tasks, \gpvone\ supports 4 and \vltf\ supports 3. 
In addition, each model is also trained with and without training data from \web. We evaluate these models on in-domain test sets for each task as well as on the \web\ and \opensce\ test sets.

% \subsection{Experimental Setups}
We now summarize the tasks and training details. See Figure~\ref{fig:model_io} for the inputs/outputs for each task and Supplementary Sec. 6 for additional experimental details.
\textbf{VQA:} We train on the \vqa~\cite{goyal2017cvpr_balanced_vqa_v2} train set and report results using the annotator-weighted metric from~\cite{goyal2017cvpr_balanced_vqa_v2} on the \vqa\ test-dev set and \opensce\ test set.
\textbf{Captioning:} We train on \coco\ captioning and report CIDEr-D~\cite{Vedantam2015CIDErCI} on \coco\ test. \opensce~uses nocaps~\cite{Agrawal2019nocapsNO} for captioning. Due to space constraints, we only report CIDEr-D on the out-of-domain split, as performance on novel concepts is our primary interest. See Supplementary Sec. 11 for results on all splits. 
\textbf{Localization:} Localization training data is built from bounding box annotations in \coco\ images following \cite{gpv1}. We report mAP on the \coco\ val set (since the test servers do not support this task) and the \opensce\ test set. \vltf\ does not support this task out-of-the-box since it does not have a means to rank its input boxes, so we do not train or evaluate it for this task.
\textbf{Classification:} We use the classification data from~\cite{gpv1} and report accuracy on the \coco\ val set and the \opensce\ test set. Since \opensce\ is out-of-domain we apply the re-calibration method from Sec.~\ref{sec:model} for \gpvtwo. 
\textbf{Classification-in-Context:} The same as classification, except instead of cropping images the bounding box of the target object is used as an input box. Having an input box means only \gpvtwo\ supports this task.

% SPACE: Could just remove the comment about batching
\oursubsection{Training details} We train \gpvtwo\ and \vltf\ for 8 epochs with a batch size of 60 and learning rate of 3e-4 that linearly warms up from 0 for 10\% of the training steps then decays to 0. We stratify the data so examples from each source are proportionally represented in each batch. Since the web data is large, we shard the data into 4 parts and use 1 shard each epoch, resulting in about a third of the data in each epoch being web data. \vltf\ is initialized with a pre-trained checkpoint~\cite{vlt5} and \gpvtwo\ is initialized from our checkpoint after \cc\ pre-training. We train \gpvone\ to 40 epochs following \cite{gpv1}\footnote{Since \cite{gpv1} takes a long time to train when using the web data (over 3 weeks), results for \gpvone\ with and without web data are reported after 20 epochs training.}.

\input{tables/tab_concept_scaling_closed_world}
\medskip
\oursubsection{Concept expansion using web data}
Table~\ref{tab:concept_scaling} shows the performance of models when trained with and without \web. On \opensce, which contains a more diverse set of concepts than \coco, we find that all models benefit from web data and perform better on captioning and the two classification tasks (with large gains of +7.1, +9.1, +8.6 for \gpvtwo). We see modest gains of +1.0 for \opensce\ localization. VQA shows small gains, presumably because many frequent answers such as colors or numbers are common between \coco\ and \opensce, and adding web supervision brings little benefits for such questions. Training with web data makes little difference on \coco\ and, unsurprisingly, leads to large gains on \web\ test, where models achieve over 40\% accuracy on nouns and 60\% on verbs despite the large number of concepts. Overall, these results show multi-tasking GPVs with web data improves performance significantly on concepts unseen in supervised data without compromising in-domain performance.

Of the three GPVs we test, we find \gpvtwo\ to be the most effective across all three benchmarks. 
\gpvtwo\ uses less pre-training data and a simpler and cheaper pre-training strategy than \vltf. However, it uses more powerful VinVL~\cite{Zhang2021VinVLMV} features and benefits from classifier re-calibration (See Tab.~\ref{tab:ablations}). In contrast to \vltf, \gpvtwo\ can also perform CiC and localization. In contrast to \gpvone, \gpvtwo\ has more shared features and a better pre-trained language model, which help produce large gains across the benchmarks. It also trains much faster than \gpvone\ as it can use pre-computed detection features (1 day on 2 GPUs vs. $>$3 weeks on 4 GPUs). See Supplementary Secs. 10 and 5 for more comparisons and \gpvtwo\ efficiency metrics respectively.
\gpvtwo\ also achieves state-of-the-art on the GRIT benchmark \cite{Gupta2022GRIT} at the time of submission (Supplementary Sec. 9).

\input{tables/tab_ablations}

% \subsection{Closed world evaluation of web data}
\oursubsection{Closed world evaluation of web data}
Table~\ref{tab:concept_scaling_closed_world} shows results for \gpvtwo\ when it is trained on the \cocosce~\cite{gpv1} dataset, a dataset that holds out different concepts from each \coco\ training supervised dataset (e.g., captions that mention the word ``bed'' are held out from the caption training data), and then evaluates whether models can still perform well on those unseen concepts by learning about them from the data in other tasks (e.g., captions with the word ``bed" are in the captioning test set, and classification and localization training still include examples about beds). When \gpvtwo\ is trained on \cocosce\ we make two notable changes: (1) We replace VinVL features with DETR~\cite{Carion2020DETR} features trained only on the \cocosce\ training categories (this avoids leaking detection information by VinVL's broad category set); and (2) We do not pre-train with \cc\ (this avoids leaking caption information from \cc's broad vocabulary). These choices severely reduce the performance of the model, but this setup serves as a closed world evaluation to determine if \gpvtwo\ can learn skills from \cocosce\ and concepts from \web. As seen in Table~\ref{tab:concept_scaling_closed_world}, training with web data shows large gains across the board in this controlled experiment. In fact, we now also see gains in the unseen categories within \cocosce.

% \gpvtwo\ uses VinVL features and is pre-trained with \cc. While immensely useful, these sources of information prevent a closed world evaluation to determine if \gpvtwo\ can learn skills from COCO and concepts from web, since VinVL and \cc contain concept information overlapping with \opensce's concept set. 

% We additionally show results for \gpvtwo trained on \cocosce\cite{gpv1}, a dataset that holds out different concepts from each COCO training supervised datasets (e.g., captions that mention the word ``bed'' are held out from the caption training data), and then evaluates whether models can still perform well on those unseen concepts my learning about those concepts from the data in other tasks (e.g., captions with the word ``bed" are in the captioning test set, and classification and localization training still include examples about beds). Following the protocols from \cite{gpv1} we avoid the use of most forms external data by not pre-training the model, and using the a DETR~\cite{Carion2020DETR} object detector trained on only the train images instead of VinVL to generate features. As a result, this model does not use an box input since DETR cannot be used to extract features from boxes DETR did not predict. We therefore did not train or evaluate on classification-in-context.

% \subsection{Ablation analysis}
% \label{sect:ablations}
\oursubsection{Ablation analysis}
We perform ablation studies on \gpvtwo. Table~\ref{tab:ablations} shows results on the validation sets.
The model that does not use LBL scores each box using a linear classifier on top of its contextualized embedding instead.
On both classification tasks and captioning, we find that web data helps with and without \cc\ pre-training, and that removing both reduces performance dramatically ($>$30 points for captioning). This shows that the two approaches are independently effective and complementary at helping models handle new concepts. This is also true to a more modest extent for localization. Using the web data for a second round of pre-training performed better than not using it, but was significantly worse than our multi-tasking framework.
Re-calibration is critical for classification, providing a gain of up to 12 points, confirming that models tend to be overly influenced by the concept distribution observed during training.
Performance on \coco~remains largely unchanged, which is expected as our design choices target performance on unseen concepts. Finally, VinVL significantly out-performs DETR, as expected given its much more extensive training regime.

%compared to the fully supervised model, demonstrating that it has learned attribute information from its data sources. The combination of \coco\ and \cc\ data proves sufficient for it to learn about attributes, and hence \web\ shows no additional gains in this 0-shot scenario. 
% COCO VQA has impressive coverage over the objects and attributes of this dataset (\todo{Put stats}), so the web data does not provide significant benefit for this task. However, the model performs quite well on the task for the zero-shot setting, in comparison to fully-supervised methods.

% Unsure if we're adding fashion/refexp, per someone's comment in the table caption
\input{figs/qual_sample}

\oursubsection{Human object interaction}
% \label{sect:hoi}
To demonstrate the flexibility of \gpvtwo, we also employ it for human-object interaction detection~\cite{ChaoHOIDet} using the two-stage procedure described in Sec.~\ref{sec:model}. 
We fine-tune \gpvtwo\ on the \hicodet\ train set for 4 epochs (see Supplementary Sec. 7 for details).
 %This task requires models to identify pairs of regions in the image, one containing a person and one an object, and then label the object and the person's relationship to that object. We apply \gpvtwo\ using the two-pass procedure described in Sec.~\ref{sec:gpv2-outputs}. 
%We fine-tune \gpvtwo\ to model to locate all people in the image from the prompt ``locate the people", and to also use the image, prompt ``What is this person interacting with", and a box around a person as input to generate either the correct object/interaction label, or a background class label ``no interaction", for each box found in the image. See the supplementary for details.
%\todo{Make this comment on specialization more specific}
%Results are shown on the \hicodet\ data in Table~\ref{tab:hoi}, \gpvtwo\ is able to get 20.6 on the full task. 
\gpvtwo\ gets an AP of 20.6 on the \hicodet\ benchmark, which is comparable to a number of other approaches (17.2~\cite{Gupta2019NoFrillsHOI}, 19.8~\cite{Ulutan2020VSGNet}, 20.8~\cite{Zhong2020PolysemyDN}, 21.8~\cite{Fang2021DIRV}). Although recent models~\cite{kim2021hotr,zou2021end,zhang2021mining} %that predict human-object-interaction in an end-to-end fashion 
show results up to 32.1 mAP~\cite{zhang2021mining}, they require highly specialized architectures requiring up to 5 output heads (e.g. for decoding human+object boxes, interaction score, and object and interaction categories), well crafted losses (e.g. Hungarian HOI instance matching objectives), and custom post-processing steps (e.g pairwise non-maximum suppression). \gpvtwo's flexibility allows us to get reasonable results by side-stepping complex model design with simple chained inference. %without a complex task-specific design.

% \subsection{Efficient skill learning}
% \label{sect:effciency}
% \subsection{Efficiency metrics}

\oursubsection{Qualitative results from \opensce\ (Figure~\ref{fig:qual_sample})}
% Qualitative results from \opensce\ are shown in Figure~\ref{fig:qual_sample}, see Supplementary for additional examples. 
Training on \web\ helps \gpvtwo\ understand rare concepts like `sari' or `gondola', which it is able to use across diverse skills. See Supplementary Sec. 1 for more examples.

\oursubsection{Novel concepts case study}
A unique advantage of using web-search is the ability to easily and cheaply access new visual concepts that are too specialized or too recent to appear in statically-collected corpora. To demonstrate this we present qualitative results on an experiment to train \gpvtwo\ to learn a number of COVID-19 related concepts.
We collect 43 terms related to COVID-19 (e.g., N95 mask, face shield, etc.) and gather a 1000-image train set with a 100-image val set using the same automatic pipeline we used to gather \web. We fine-tune \gpvtwo\ (after it has been trained on \coco\ and \web) on these examples mixed with a sample of 2000 examples from each \coco\ train set for 3 epochs.

After fine-tuning, the model achieves 71\% accuracy on the new val set compared to only 4\% without fine-tuning (performance is initially low since these concepts are too specialized and new to appear in \cc, \coco\ or \web). See some qualitative results in Figure~\ref{fig:covid_sample} that show that \gpvtwo\ is able to use such recently-introduced concepts when applying multiple skills. Although this is a small-scale qualitative study, it shows that our approach of combining a GPV and web-search data can lead to models that not only understand a wide range of concepts and skills, but can also be efficiently adapted to new visual concepts that become common in the world or that are needed due to the specialized needs of a user. 
% Kind of struggling with this last sentence, should it citing something? Do we even need it?
We think this is an exciting avenue for future work in GPVs.

% See Supp. for qualitative results for \gpvtwo\ on the three benchmarks. \todo{Refer to figure}
\input{figs/covid_sample}

%% file: tables/tab_concept_scaling.tex
\begin{table*}[t]
\setlength{\tabcolsep}{2pt}
\footnotesize
\centering
\caption{
\label{tab:concept_scaling}
\textbf{Concept expansion with web data.} 
Jointly training on \web\ $+$ \coco\ shows consistent gains on \opensce\ and \web\ benchmarks without adversely affecting \coco\ performance for 3 different GPVs. 
% \opensce~Cap only shows out-of-domain results from nocaps due to limited space.
% Due to limited space we only show out-of-domain (OOD) performance on nocaps for \opensce~Cap. 
$\gpvone^{20}$ refers to 20 epoch training.}		
\resizebox{\textwidth}{!}{
\begin{tabular}{
    lc
    >{\columncolor{Color1}}c>{\columncolor{Color1b}}c>{\columncolor{Color1}}c>{\columncolor{Color1b}}c>{\columncolor{Color1}}c
    >{\columncolor{Color2}}c>{\columncolor{Color2b}}c>{\columncolor{Color2}}c>{\columncolor{Color2b}}c>{\columncolor{Color2}}c
    >{\columncolor{Color3}}c>{\columncolor{Color3b}}c>{\columncolor{Color3}}c>{\columncolor{Color3b}}c}

\toprule
\multicolumn{1}{l}{} & \multicolumn{1}{c}{} 
    & \multicolumn{5}{c}{\cellcolor{Color1}\coco} 
    & \multicolumn{5}{c}{\cellcolor{Color2}\opensce} 
    & \multicolumn{4}{c}{\cellcolor{Color3}\web} \\

\multicolumn{1}{c}{Model} & \multicolumn{1}{c}{Web data} 
    & \cellcolor{Color1}\emph{VQA}  
    & \cellcolor{Color1}\emph{Cap} 
    & \cellcolor{Color1}\emph{Loc}
    & \cellcolor{Color1}\emph{Cls}
    & \cellcolor{Color1}\emph{CiC}
    
    & \cellcolor{Color2}\emph{VQA}  
    & \cellcolor{Color2}\emph{Cap} 
    & \cellcolor{Color2}\emph{Loc}
    & \cellcolor{Color2}\emph{Cls}
    & \cellcolor{Color2}\emph{CiC}
    
    & \cellcolor{Color3}\emph{All}
    & \cellcolor{Color3}\emph{Nouns}
    & \cellcolor{Color3}\emph{Verbs} 
    & \cellcolor{Color3}\emph{Adj} \\
    
\midrule
\textcolor{tabindex}{[a]} \gpvone & no web & 62.5 & 102.3 & \textbf{73.0} & \textbf{83.6} & - & 45.3 & 25.8 & 61.9 & 10.1 & - & 11.9 & 2.7 & 8.5 & 24.5\\
\textcolor{tabindex}{[b]} $\gpvone^{20}$ & no web & 61.2 & 95.7 & 65.3 & 82.3 & - & 44.3 & 23.1 & 60.3 & 9.3 & - & 13.1 & 3.1 & 7.7 & 28.4\\
\textcolor{tabindex}{[c]} $\gpvone^{20}$ & with web & 61.5 & 97.3 & 64.9 & 82.8 & - & 45.8 & 28.6 & 61.5 & 20.0 & - & 54.4 & 32.7 & 51.7 & 78.8\\
\midrule
\textcolor{tabindex}{[d]} \vltf & no web & 69.8 & 100.7 & - & 78.1 & - & 60.2 & 31.6 & - & 10.9 & - & 18.6 & 4.3 & 15.8 & 35.7\\
\textcolor{tabindex}{[e]} \vltf & with web & 69.9 & 106.4 & - & 77.3 & - & 59.9 & 45.0 & - & 16.2 & - & 61.0 & 38.0 & 59.3 & \textbf{85.8}\\
\midrule
\textcolor{tabindex}{[f]} \gpvtwo & no web & 71.1 & 112.1 & 70.9 & 82.2 & \textbf{93.4} & 60.6 & 65.4 & 74.8 & 36.3 & 43.6 & 22.5 & 3.8 & 23.6 & 39.9\\
\textcolor{tabindex}{[g]} \gpvtwo & with web & \textbf{71.4} & \textbf{113.0} & 70.9 & 82.3 & 93.2 & \textbf{61.1} & \textbf{72.5} & \textbf{75.9} & \textbf{45.4} & \textbf{52.2} & \textbf{62.0} & \textbf{41.7} & \textbf{60.0} & 84.3\\
\bottomrule
\end{tabular}}
\end{table*}

%% file: tables/tab_concept_scaling_closed_world.tex
\begin{table*}[t]
\setlength{\tabcolsep}{2pt}
\footnotesize
\centering
\caption{
\label{tab:concept_scaling_closed_world}
\textbf{Concept scaling using web data: Closed world experiment.} To eliminate the effect of VinVL features and \cc\ pretraining, we restrict \gpvtwo\ to \cocosce\ trained DETR features. Training jointly with \web\ still shows massive gains on \opensce\ and \web\ vs training with only \cocosce.
}		
% \begin{subtable}[h]{1.0\textwidth}\centering
\resizebox{\textwidth}{!}{
\begin{tabular}{
    lc
    >{\columncolor{Color4}}c>{\columncolor{Color4}}c>{\columncolor{Color4}}c
    >{\columncolor{Color4b}}c>{\columncolor{Color4b}}c>{\columncolor{Color4b}}c
    >{\columncolor{Color4}}c>{\columncolor{Color4}}c>{\columncolor{Color4}}c
    >{\columncolor{Color4b}}c>{\columncolor{Color4b}}c>{\columncolor{Color4b}}c
    >{\columncolor{Color2}}c
    >{\columncolor{Color2b}}c
    >{\columncolor{Color2}}c>{\columncolor{Color2b}}c
    >{\columncolor{Color3}}c>{\columncolor{Color3b}}c>{\columncolor{Color3}}c>{\columncolor{Color3b}}c
    }

\toprule
\multicolumn{2}{c}{} 
    & \multicolumn{12}{c}{\cellcolor{Color4}\cocosce}
    & \multicolumn{4}{c}{\cellcolor{Color2}\opensce}
    & \multicolumn{4}{c}{\cellcolor{Color3}\web} \\

\multicolumn{2}{c}{} 
    & \multicolumn{3}{c}{\cellcolor{Color4}\emph{VQA}}
    & \multicolumn{3}{c}{\cellcolor{Color4}\emph{Cap}} 
    & \multicolumn{3}{c}{\cellcolor{Color4}\emph{Loc}} 
    & \multicolumn{3}{c}{\cellcolor{Color4}\emph{Cls}}
    % & \multicolumn{3}{c}{\cellcolor{Color4}\emph{CiC}} \\
    & \multicolumn{1}{c}{\cellcolor{Color2}\emph{VQA}}
    & \multicolumn{1}{c}{\cellcolor{Color2}\emph{Cap}} 
    & \multicolumn{1}{c}{\cellcolor{Color2}\emph{Loc}} 
    & \multicolumn{1}{c}{\cellcolor{Color2}\emph{Cls}} 
    % & \multicolumn{1}{c}{\cellcolor{Color2}\emph{CiC}} 
    
    & \multicolumn{1}{c}{\cellcolor{Color3}\emph{All}} 
    & \multicolumn{1}{c}{\cellcolor{Color3}\emph{Noun}} 
    & \multicolumn{1}{c}{\cellcolor{Color3}\emph{Verb}} 
    & \multicolumn{1}{c}{\cellcolor{Color3}\emph{Adj}} \\

\multicolumn{1}{c}{Model} & \multicolumn{1}{c}{Web data} 
    & \cellcolor{Color4}\emph{Test} & \cellcolor{Color4}\emph{Sn} & \cellcolor{Color4}\emph{Unsn}
    & \cellcolor{Color4}\emph{Test} & \cellcolor{Color4}\emph{Sn} & \cellcolor{Color4}\emph{Unsn}
    & \cellcolor{Color4}\emph{Test} & \cellcolor{Color4}\emph{Sn} & \cellcolor{Color4}\emph{Unsn}
    & \cellcolor{Color4}\emph{Test} & \cellcolor{Color4}\emph{Sn} & \cellcolor{Color4}\emph{Unsn}
    % & \cellcolor{Color4}\emph{Test} & \cellcolor{Color4}\emph{Sn} & \cellcolor{Color4}\emph{Unsn} \\
    & \cellcolor{Color2}{} 
    & \cellcolor{Color2}{} 
    % & \cellcolor{Color2}\emph{In} & \cellcolor{Color2}\emph{Near} & \cellcolor{Color2}\emph{Out} & \cellcolor{Color2}\emph{All}
    & \cellcolor{Color2}{} & \cellcolor{Color2}{} 
    
    & \cellcolor{Color3}{} & \cellcolor{Color3}{} & \cellcolor{Color3}{} & \cellcolor{Color3}{}\\
    
\midrule
\gpvtwo & no web &  59.6 & 60.1 & 48.5 & 88.4 & 91.7 & 55.5 & 62.2 & \textbf{67.2} & 14.0 & \textbf{73.1} & 77.2 & \textbf{33.9} & \textbf{46.9} & 21.1 & 54.9 & 13.6 & 14.0 & 3.3 & 11.6 & 27.1 \\
\gpvtwo & with web & \textbf{59.9} & \textbf{60.3} & \textbf{49.7} & \textbf{89.2} & \textbf{92.1} & \textbf{58.0} & 62.2 & 67.0 & \textbf{14.8} & 73.0 & 77.2 & 32.6 & 46.8 & \textbf{33.4} & \textbf{58.7} & \textbf{26.5} & \textbf{47.0} & \textbf{25.1} & \textbf{43.0} & \textbf{73.0} \\
\bottomrule
\end{tabular}}
\end{table*}

%% file: tables/tab_ablations.tex
\begin{table*}[t]
\setlength{\tabcolsep}{5pt}
\footnotesize
\centering
\caption{
\label{tab:ablations}
\textbf{Ablating \gpvtwo}. The left-most columns indicate using \web~(`Pre.' indicates pre-training with \web\ instead of multi-tasking), \cc\ pre-training, classifier re-calibration (Cb), language-based localization (LBL) (see Sec.~\ref{sec:model}), and VinVL instead of the DETR detector from \gpvone. The first row shows results for \gpvtwo, and the lower rows show the differences in scores between ablations and \gpvtwo. Each component improves performance on \opensce.
% \opensce\ Cap shows out-of-domain results.
}
\addtolength{\tabcolsep}{-1pt}
\resizebox{\textwidth}{!}{
\begin{tabular}{
    ccccc
    >{\columncolor{Color1}}c>{\columncolor{Color1b}}c>{\columncolor{Color1}}c>{\columncolor{Color1b}}c>{\columncolor{Color1}}c
    >{\columncolor{Color2}}c>{\columncolor{Color2b}}c>{\columncolor{Color2}}c>{\columncolor{Color2b}}c>{\columncolor{Color2}}c
    >{\columncolor{Color3}}c>{\columncolor{Color3b}}c>{\columncolor{Color3}}c>{\columncolor{Color3b}}c}

\toprule
\multicolumn{1}{c}{} & \multicolumn{1}{c}{} & \multicolumn{1}{c}{} & \multicolumn{1}{c}{} & \multicolumn{1}{c}{}
    & \multicolumn{5}{c}{\cellcolor{Color1}\coco} 
    & \multicolumn{5}{c}{\cellcolor{Color2}\opensce} 
    & \multicolumn{4}{c}{\cellcolor{Color3}\web} \\

Web & CC & Cb & LBL & Vin.
    & \cellcolor{Color1}\emph{VQA}  
    & \cellcolor{Color1}\emph{Cap} 
    & \cellcolor{Color1}\emph{Loc}
    & \cellcolor{Color1}\emph{Cls}
    & \cellcolor{Color1}\emph{CiC}
    
    & \cellcolor{Color2}\emph{VQA}  
    & \cellcolor{Color2}\emph{Cap} 
    & \cellcolor{Color2}\emph{Loc}
    & \cellcolor{Color2}\emph{Cls}
    & \cellcolor{Color2}\emph{CiC}
    
    & \cellcolor{Color3}\emph{All}
    & \cellcolor{Color3}\emph{Nouns}
    & \cellcolor{Color3}\emph{Verbs} 
    & \cellcolor{Color3}\emph{Adj} \\
    
\midrule
\checkmark & \checkmark & \checkmark & \checkmark & \checkmark & 70.7 & 117.3 & 70.9 & 82.3 & 93.2 & 60.7 & 78.0 & 76.8 & 45.8 & 52.2 & 60.4 & 39.9 & 57.5 & 83.8\\
- & \checkmark & \checkmark & \checkmark & \checkmark & -0.2 & -1.1 & 0.0 & -0.1 & 0.2 & -0.5 & -8.8 & -1.0 & -8.5 & -7.4 & -37.2 & -35.4 & -32.5 & -43.8\\
Pre. & \checkmark & \checkmark & \checkmark & \checkmark & -0.4 & -0.6 & 0.0 & -0.2 & 0.1 & -0.8 & -8.2 & -1.3 & -6.2 & -5.5 & -31.3 & -30.4 & -27.6 & -35.9\\
\checkmark & - & \checkmark & \checkmark & \checkmark & 0.4 & -2.4 & 0.1 & 0.5 & 0.1 & 0.8 & -13.9 & -0.7 & -4.3 & -4.5 & -2.3 & -3.7 & -2.4 & -0.9\\
- & - & \checkmark & \checkmark & \checkmark & 0.2 & -4.2 & 0.1 & 0.5 & 0.2 & -0.2 & -33.7 & -4.5 & -20.7 & -21.1 & -40.6 & -37.4 & -39.3 & -44.9\\
\checkmark & \checkmark & - & \checkmark & \checkmark & 0.0 & 0.0 & 0.0 & 0.0 & 0.0 & 0.0 & 0.0 & 0.0 & -11.8 & -12.8 & 0.0 & 0.0 & 0.0 & 0.0\\
\checkmark & \checkmark & \checkmark & - & \checkmark & -0.1 & -1.4 & 0.0 & 0.3 & 0.0 & -0.2 & -2.4 & -1.3 & -1.3 & -0.7 & 0.1 & 0.2 & 0.7 & -0.8\\
\checkmark & \checkmark & \checkmark & \checkmark & - & -8.1 & -15.8 & 6.1 & -2.2 & - & -9.8 & -41.7 & -15.0 & -17.4 & - & -13.8 & -13.3 & -16.4 & -11.7\\

\bottomrule
\end{tabular}}
\end{table*}

%% file: figs/qual_sample.tex
\begin{figure}[t]
    \centering
    \includegraphics[width=\columnwidth]{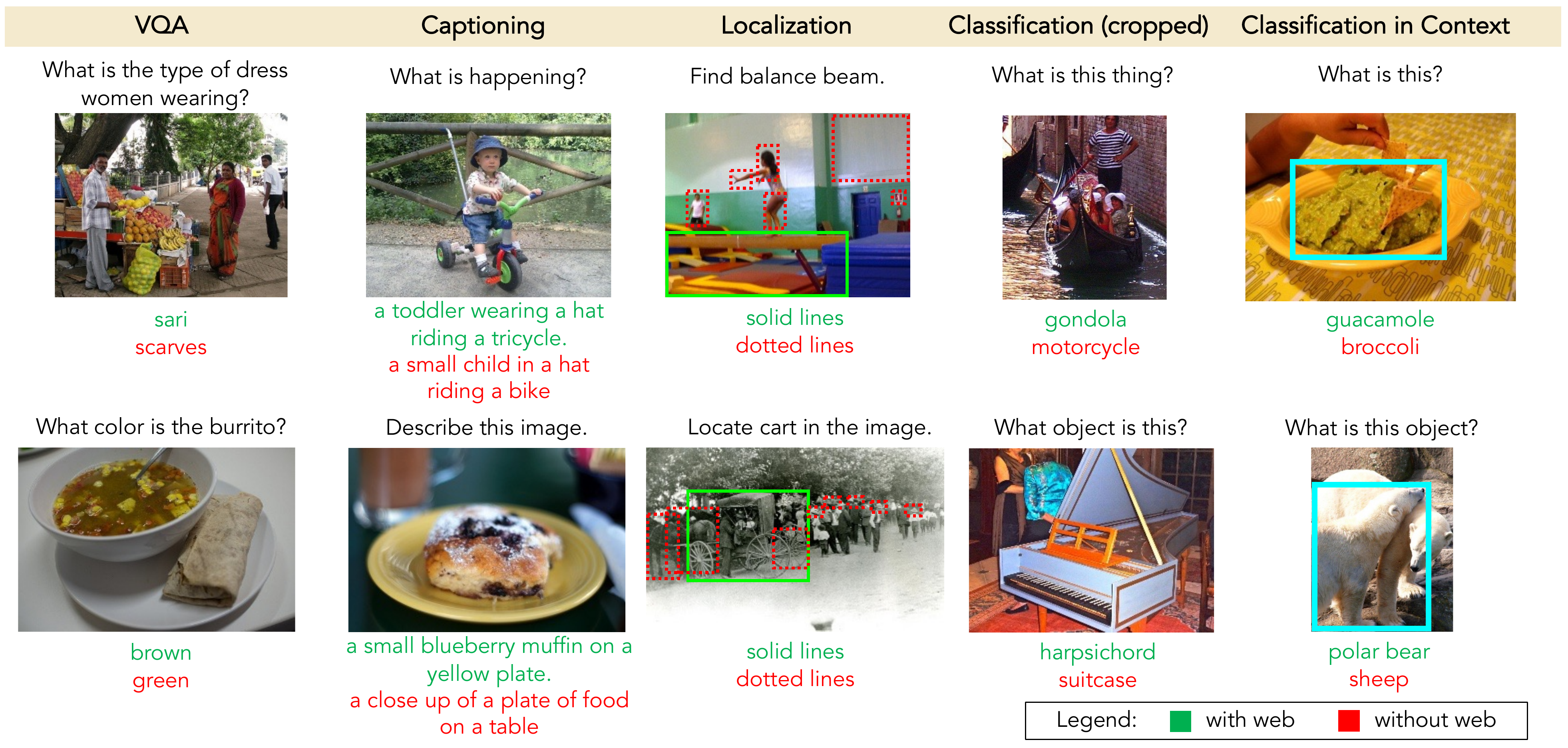}
    \caption{\textbf{Qualitative results of \gpvtwo\ on \opensce\ with and without WEB10K:} 
    %Input images, bounding boxes, and prompts are shown with the outputs of \gpvtwo~and \gpvtwo~without web training below. 
    Without web training, \gpvtwo~can ignore concepts rarely seen in the supervised training data (e.g., `balance beam' top middle) or predict frequently occurring concepts that do not appear in the image (e.g., `sheep' lower right). Web training fixes these issues and allows generalization to rare concepts like `sari' and `harpsichord'.}
    \label{fig:qual_sample}
    \vspace{-0.2in}
\end{figure}

%% file: figs/covid_sample.tex
\begin{figure}[t]
    \centering
    \includegraphics[width=\columnwidth]{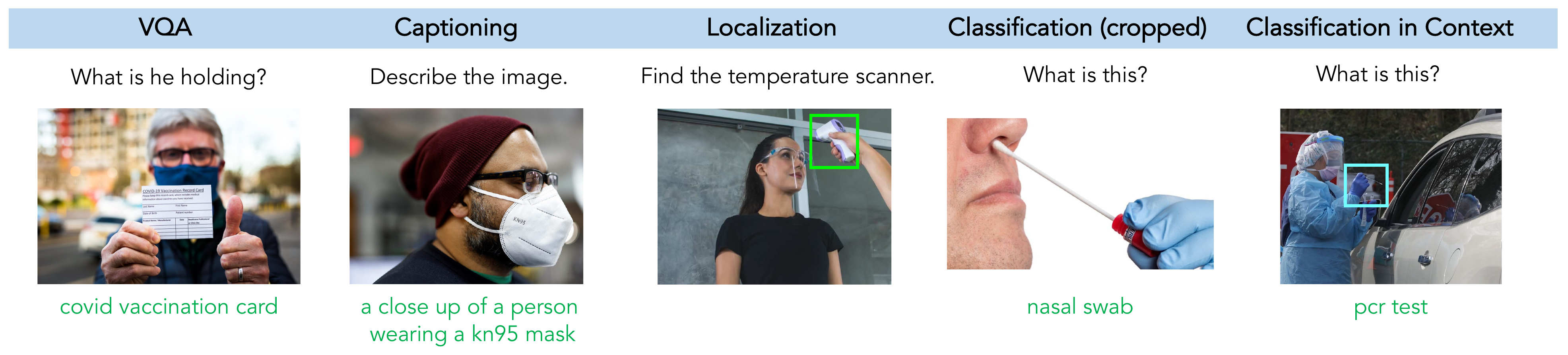}
    \caption{\textbf{Qualitative results on novel concepts:} The predictions of \gpvtwo\ after fine-tuning on COVID-related web data. The model can recognize the new concepts in new images across all skills after training on only $\sim$20 images per concept.}
    \label{fig:covid_sample}
    \vspace{-0.15in}
\end{figure}

%% file: sections/7_discussion.tex
%\vspace{-0.5in}
\section{Discussion}
\oursubsection{Extensions}
\gpvtwo\ achieves transfer of concepts from web data to skills, but our results indicate that more work is needed, particularly for tasks like VQA or localization, through new architectures or training protocols.
% \gpvtwo\ supports a wide range of tasks, but handling additional modalities (e.g., video, segmentation, such as in~\cite{perceiverIO}) and raises the potential of transferring concepts from web data to an even wider set of tasks.
\gpvtwo\ supports many tasks, but could be extended to handle more modalities (e.g., video) and outputs (e.g., segmentation). Recent work shows promise in this regard~\cite{perceiverIO}, potentially enabling transfer of web concepts to a wider range of tasks.

% Though our classifier re-calibration (Sec.~\ref{sect:classification-calibration}) is effective at improving classification transfer, it requires using an out-of-domain validation set for tuning which might not always be available, so a more broadly applicable method would be useful. 
%More work is needed to transfer concepts effectively and efficiently. \gpvtwo~shows promise but our numbers reflect that more work is needed, particularly to transfer to VQA or localization. This could include more architectural exploration in line with the ideas we propose in Sec.~\ref{sect:gpv2-outputs}, or using training protocols design specifically to improve concept transfer. Additionally, while classifier re-calibration (Sec.~\ref{sect:classification-calibration}) was very effective at improving classification transfer, it requires using out-of-domain validation set for tuning which might not always be available, so a more broadly applicable method would be useful.

%\oursubsection{Potential negative impact}
%\label{sec:negative}
% In this we work we train models that can perform vision and vision+language tasks across a more diverse set of concepts. This can potentially empower malicious uses of these models, for example using such models to profile people by analyzing their photographs or to impersonate humans. 

\oursubsection{Conclusion} As the vision community builds progressively more general models, identifying efficient ways of learning a large variety of skills and concepts is of prime importance. Our work revisits the idea of webly-supervised learning in the context of GPVs and shows that learning skills from task-datasets and concepts from the web is an efficient and inexpensive option for concept expansion.

\oursubsection{Acknowledgements} This work is partially supported by ONR award N00014-21-1-2705.
%\todo{Add a conclusion}

%% file: sections/11_appendix.tex
\setcounter{figure}{0}
\setcounter{table}{0}
\setcounter{footnote}{0}
\setcounter{page}{1}
\setcounter{section}{0}

\title{Webly Supervised Concept Expansion for General Purpose Vision Models\\{\small Supplementary Material}}

\titlerunning{Webly Supervised Concept Expansion for GPVs: Supplementary}
% If the paper title is too long for the running head, you can set
% an abbreviated paper title here
%
\author{Amita Kamath\thanks{Equal contribution}\inst{1} \and
Christopher Clark\printfnsymbol{1}\inst{1} \and
Tanmay Gupta\printfnsymbol{1}\inst{1} \and
Eric Kolve\inst{1} \and
Derek Hoiem\inst{2} \and
Aniruddha Kembhavi\inst{1}}
\authorrunning{A. Kamath et al.}
% First names are abbreviated in the running head.
% If there are more than two authors, 'et al.' is used.
%
\institute{Allen Institute for Artificial Intelligence \and
University of Illinois at Urbana-Champaign
%\email{\{amitak, chrisc, tanmayg, erick, anik\}@allenai.org}\\
%\email{dhoiem@illinois.edu}
}

%******************

\maketitle

The supplementary includes the following sections:
\begin{itemize}
\itemsep0em 
    \item[$\bullet$] Sec~\ref{app:qual}: Qualitative results from \gpvtwo
    \item[$\bullet$] Sec~\ref{app:recalibration}: Classification re-calibration analysis
    \item[$\bullet$] Sec~\ref{app:web-questions}: \web~questions and statistics
    \item[$\bullet$] Sec~\ref{app:opensce-sampling}: \opensce~sampling details
    \item[$\bullet$] Sec~\ref{app:efficiency}: \gpvtwo~efficiency metrics
    % \item Sec~\ref{app:dce_specialized} - Specialized models on \opensce
    \item[$\bullet$] Sec~\ref{app:training_details}: Experimental details
    \item[$\bullet$] Sec~\ref{app:hoi_details}: Human Object Interaction experimental details
    \item[$\bullet$] Sec~\ref{app:verbs_attr}: Zero-shot verb and attribute recognition
    \item[$\bullet$] Sec~\ref{app:grit_numbers}: Performance on the GRIT benchmark
    \item[$\bullet$] Sec~\ref{app:direct_comparison}: Comparison between the \gpvtwo\ and \gpvone\ architectures when trained on the same data
    \item[$\bullet$] Sec~\ref{app:nocaps_full_results}: Results on all nocaps splits for \opensce\ captioning
    \item[$\bullet$] Sec~\ref{app:neg_impact}: Biases in web data
    % \item Sec~\ref{app:license} - License information
\end{itemize}

\section{Qualitative results from \gpvtwo}
\label{app:qual}

\begin{figure*}
    \centering
    \includegraphics[width=0.9\textwidth]{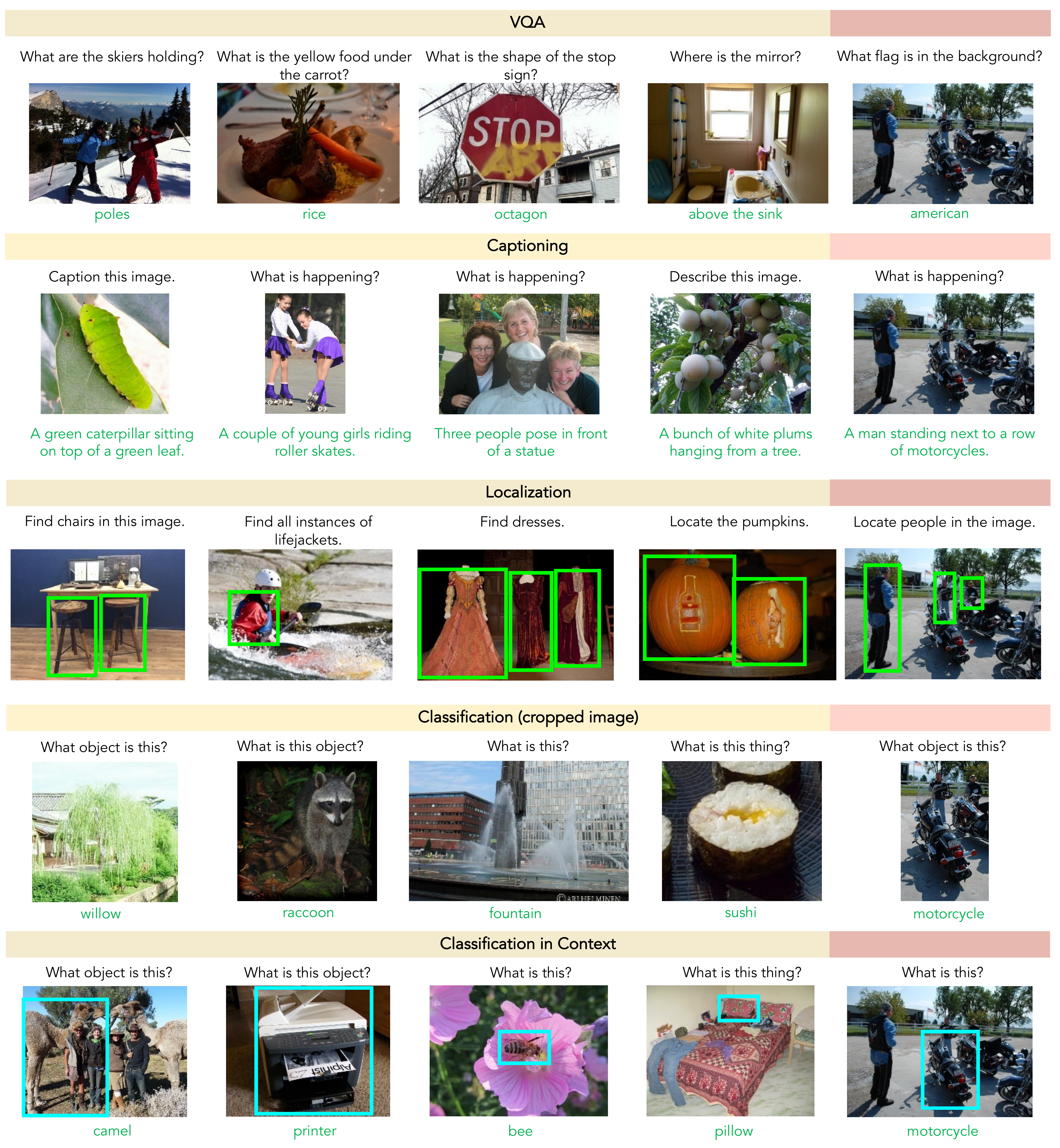}
    \caption{\textbf{Qualitative examples for \gpvtwo}. Examples are from \opensce\ val, except for the last image in each row, which comes from \coco\ val. \gpvtwo\ is able to use concepts that do not appear in the \coco\ training data across all five skills.}
    \label{fig:qualitative}
\end{figure*}
\begin{figure*}
    \centering
    \includegraphics[width=0.95\textwidth]{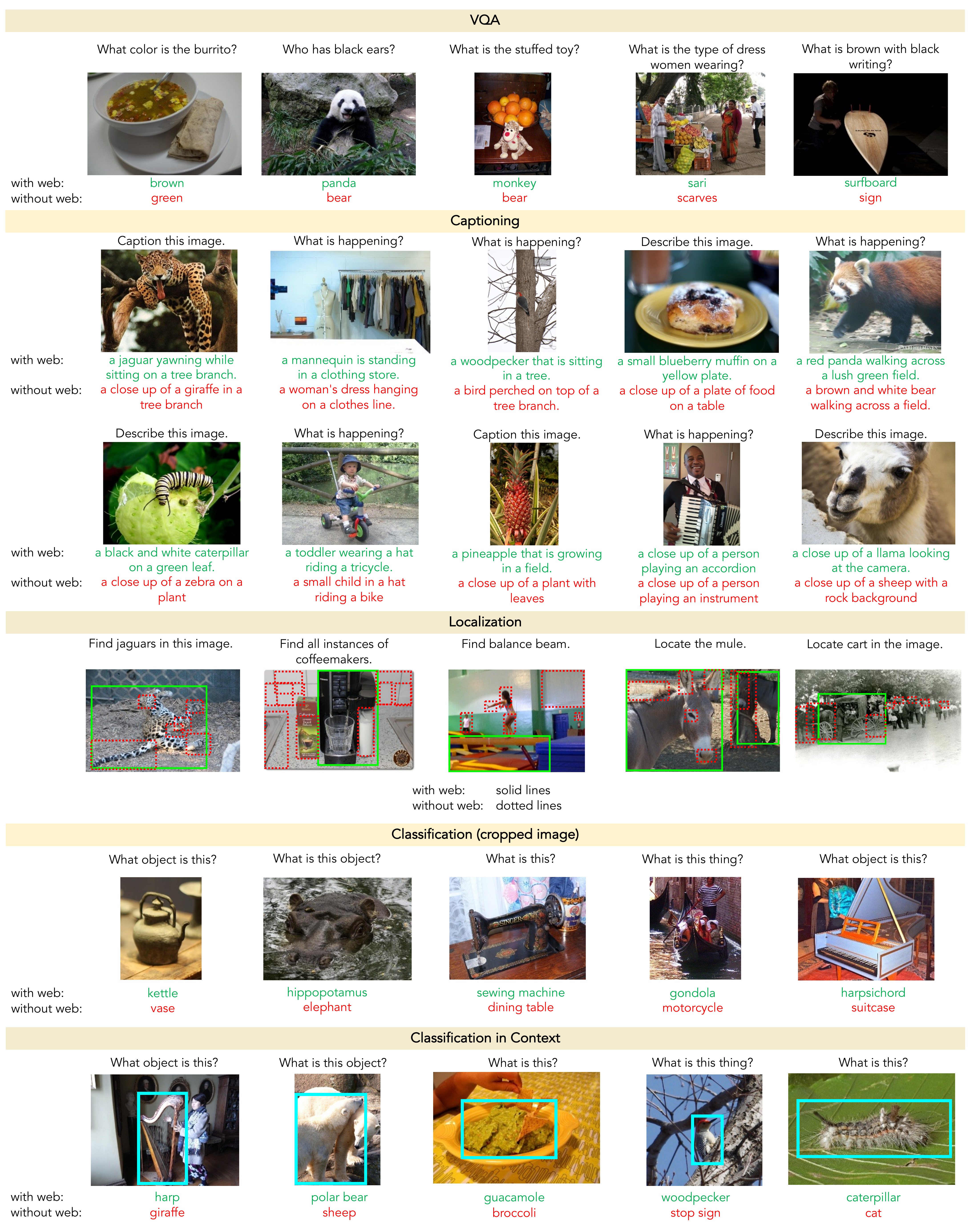}
    \caption{\textbf{Qualitative Examples: \gpvtwo~on \opensce, with and without training on WEB10K}. The use of \web~allows \gpvtwo~to understand more concepts across all skills, especially for rare concepts such as ``red panda'' (captioning upper right).}
    \label{fig:web_nonweb}
\end{figure*}

Qualitative results from \gpvtwo\ are shown in Figure~\ref{fig:qualitative}. Despite the presence of concepts that are not annotated in \coco\ (e.g, ``Caterpillar'', ``Lifejackets", ``Willow") \gpvtwo\ is able to successfully perform classification, localization, captioning, and visual questioning answering. Visualizations of predictions from \gpvtwo\ on \textit{randomly selected} examples from the \coco, \opensce, and \web\ datasets can be found in additional files in the supplementary materials. 

Figure~\ref{fig:web_nonweb} contains an expanded version of Figure 4 from the paper showing the predictions of \gpvtwo\ when trained with and without \web. The model trained without web data generates \coco\ concepts even when they are not present in the image (e.g., writing a caption about a giraffe for a picture of a jaguar, a brown-and-white bear for a red panda, or classifying a monkey as a bear), while the model trained on web data is able to name the new concepts correctly. For localization, we observe cases where the model trained without \web\ struggles on new concepts (e.g., the without web model focuses on cups or the background for the class ``coffemaker") while the model trained with \web\ can localize them accurately.

\section{Classification re-calibration analysis}
\begin{figure}
    \centering
    \includegraphics[width=0.6\textwidth]{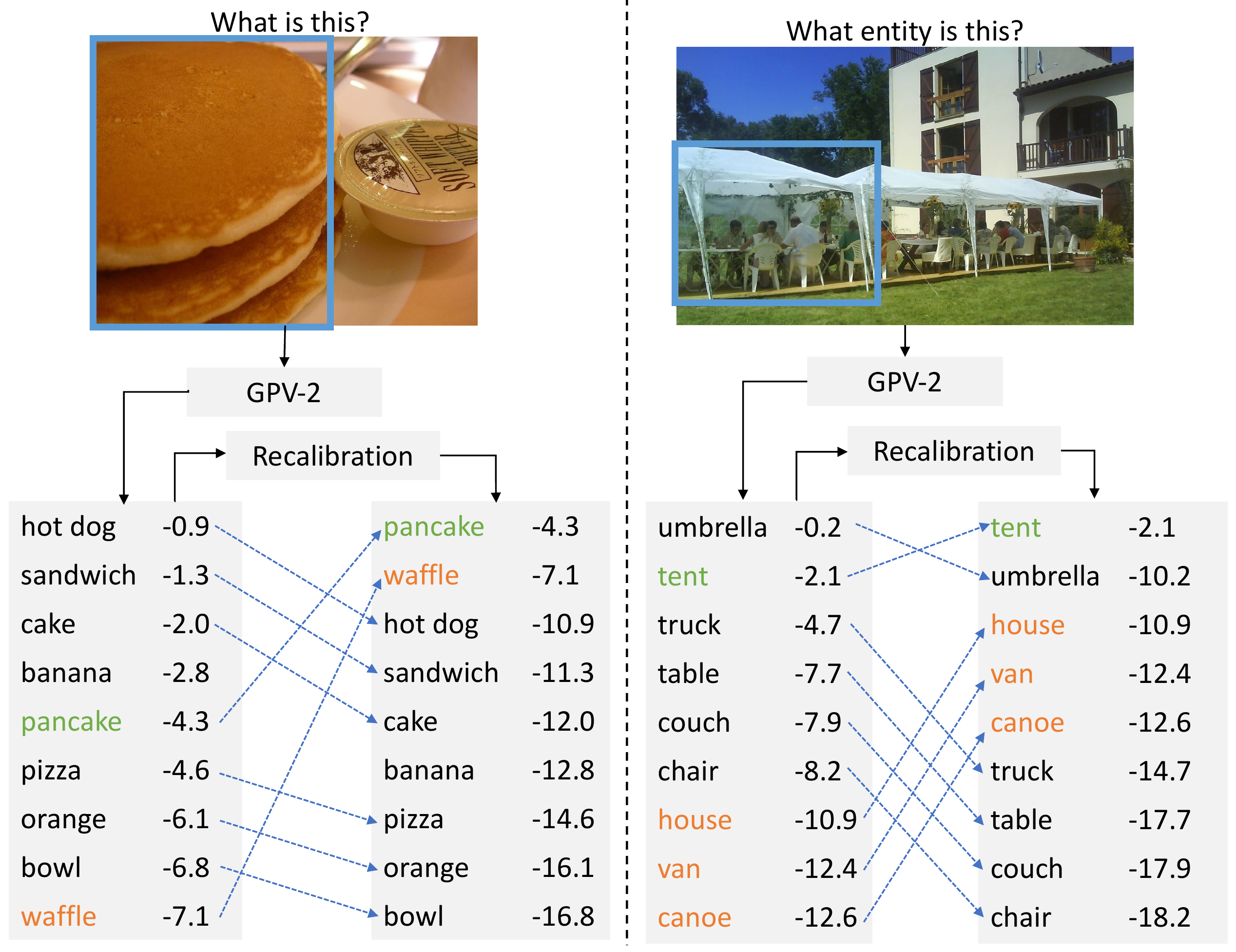}
    \caption{\textbf{Qualitative examples of re-calibration}. This figure shows two CiC examples, where the left tables show \gpvtwo's top 9 predictions and log-probability scores, and the right table shows how the scores and rankings change after re-calibration. The model has a strong preference for answers seen in the \coco\ classification data (black), resulting in the model ranking \coco\ classes that are vaguely visually similar to the image over the correct class (green). Re-calibration increases the relative score of the non-\coco\ answers (green if correct, orange otherwise) allowing the model to get these examples correct.}
    \label{fig:recalibration}
\end{figure}

\label{app:recalibration}
In this section, we analyze the classification re-calibration method from Sec.~\ref{sec:model}. 
Table~\ref{tab:recalibration} shows a breakdown of how \gpvtwo\ behaves on \opensce\ classification with and without re-calibration. 
Without re-calibration \gpvtwo\ predicts a \coco\ category for 56\% of CiC examples and 65.7\% of the CLS examples, even though only 14\% of these examples belong to a \coco\ category, showing that the model has a strong bias towards these categories. Adding re-calibration mostly mitigates this bias and significantly boosts performance on non-\coco\ categories. It comes at the cost of some performance on examples that belong to \coco\ categories, but those examples are only a small portion of the data so performance is increased by 12 points overall. These results show re-calibration is an important component to allowing models to transfer concepts learned from non-classification data to the classification skill. Qualitative examples are shown in Figure~\ref{fig:recalibration}. 

\begin{table}[h!]
    \footnotesize
    \centering
    \caption{\textbf{\gpvtwo\ accuracy on \opensce\ classification with and without classifier re-calibration (Cb)}. The Acc. column shows overall accuracy, \coco\ Acc. shows accuracy on examples with labels in the 80 \coco\ categories, Other Acc. shows accuracy on other examples, and \coco\ Ans. shows how often the model predicts a \coco\ category.}
    \label{tab:recalibration}
    \begin{tabular}{c|c|c|c|c|c} 
    \toprule
         Task & Cb & Acc. & \coco\ Acc. & Other Acc. & \coco\ Ans. \\ 
         \midrule
         CiC & - & 39.4 & 92.0 & 30.8 & 56.4 \\
         CiC & \checkmark & 52.2 & 77.5 & 48.1 & 19.7 \\
         CLS & - & 34.0 & 85.7 & 25.5 & 65.7 \\
         CLS & \checkmark & 45.8 & 69.9 & 41.9 & 24.2 \\
         \bottomrule
         
    \end{tabular} 
    %\vspace{0.05in} %%
    
\end{table}

\section{WEB10K questions and statistics}
In this section, we provide more detail about how we construct question-answer pairs from the web search data. 
For each query-image pair, we construct a question that is answered by the noun from the query. For example, the question ``What entity is this?" with the answer ``dog" for the query ``brown dog". For queries that contain a verb, we construct two additional questions that are answered by the verb, one that specifies the noun and one that does not. For example, ``What action is happening?", and ``What is the dog doing?" with the answer ``running", for the query ``dog running". For queries that contain adjectives, we similarly construct two questions that are answered by the adjective, one that specifies the noun and one that does not. To do this, we manually map the adjectives to adjective types (e.g., ``color" for ``red") and specify the adjective type in the question. For example, ``What is the color of this object?" and ``What is the color of this dog?" with the answer ``brown", for the query ``brown dog". Using adjective types is important to because generic questions like ``What attributes does this object have?" will have many possible correct answers. 
Finally, for all query-image pairs, we additionally construct a query whose answer is the entire query. 
During evaluation, we compute the average accuracy on questions where the is answer is a noun, verb or adjective, and report the macro-average of those results to get an overall accuracy number.

The questions themselves are generated by a templating system to increase their linguistic diversity. Table~\ref{tab:webqa-templates} shows the templates we use. For a given query and question type we use these templates to generate a large number of possible questions, and then select one at random to use as a prompt for the model.

Additional question types are possible. For example, contrastive questions like ``Is this sloth swimming or climbing?", or questions that specify hypernyms of the answer (obtained from sources such as WordNet) like ``What kind of reptile is this?".
We leave the generation of such questions, as well as their impact on knowledge transfer of concepts between skills, to future work.

\begin{table}[]
\footnotesize
    \centering
    \caption{\textbf{Templates for generating web prompts}. Templates are grouped by whether they have a noun, verb, or adjective answer. These templates are expanded by substituting the all-caps words for any one of the substitute words specified below the table, except ADJ\_TYPE which is replaced by the type of the adjective for questions with adjective answers. For verb and adjective questions where the object is specified, OBJ is replaced by the noun instead, and verb templates that do not contain OBJ are not used. \\
    }
    \begin{tabular}{c|l}
        \toprule
        \textbf{Answer Type} & \textbf{Prompts} \\
        \midrule
      \multirow{6}{*}{Noun} &  What is DT OBJ? \\
      & What OBJ is this? \\
      & What OBJ is that? \\
      & Classify DT OBJ. \\
      & Specify DT OBJ. \\
      & Name DT OBJ. \\
      \midrule
      \multirow{3}{*}{Adjective} &  WH ADJ\_TYPE is DT OBJ? \\
       & What is the ADJ\_TYPE of DT OBJ? \\
       & CMD the ADJ\_TYPE of DT OBJ.\\
       \midrule
      \multirow{17}{*}{Verb} & What is DT OBJ doing? \\
& What action is DT OBJ taking? \\
& What action is DT OBJ performing? \\
& What action is DT OBJ carrying out? \\
& What action is DT OBJ doing? \\
& What activity is DT OBJ doing? \\
& CMD the action being taken by DT OBJ. \\
& CMD the activity DT OBJ is doing. \\
& CMD what DT OBJ is doing. \\
      & What is being done? \\
      & WH action is being done? \\
      & WH activity is being done? \\
      & WH activity is this? \\
      & WH action is being taken? \\
      & CMD the activity being done. \\
      & CMD the action being done. \\
      & CMD the action being taken. \\
      & What is DT OBJ doing? \\
      \midrule
      \multirow{2}{*}{Entire Query} & What is this? \\
      & What is that? \\
        \bottomrule

    \end{tabular}
    
    % \begin{tabular}{c|c}
    %\vspace{0.2cm} %%
    \raggedright
    %\vskip 1em
    ~\\
     DT $\rightarrow$ the, this, that  \\
     OBJ $\rightarrow$ entity, object  \\
     WH $\rightarrow$ What, Which \\
     CMD $\rightarrow$ Describe, State, Specify, Name \\
    % \end{tabular}
    %\vspace{0.05in}
    \label{tab:webqa-templates}
\end{table}
\label{app:web-questions}

\section{\opensce~sampling details}
\label{app:opensce-sampling}

\begin{figure}[ht]
\begin{subfigure}{\linewidth}
  \centering
  % include first image
  \includegraphics[width=0.7\linewidth]{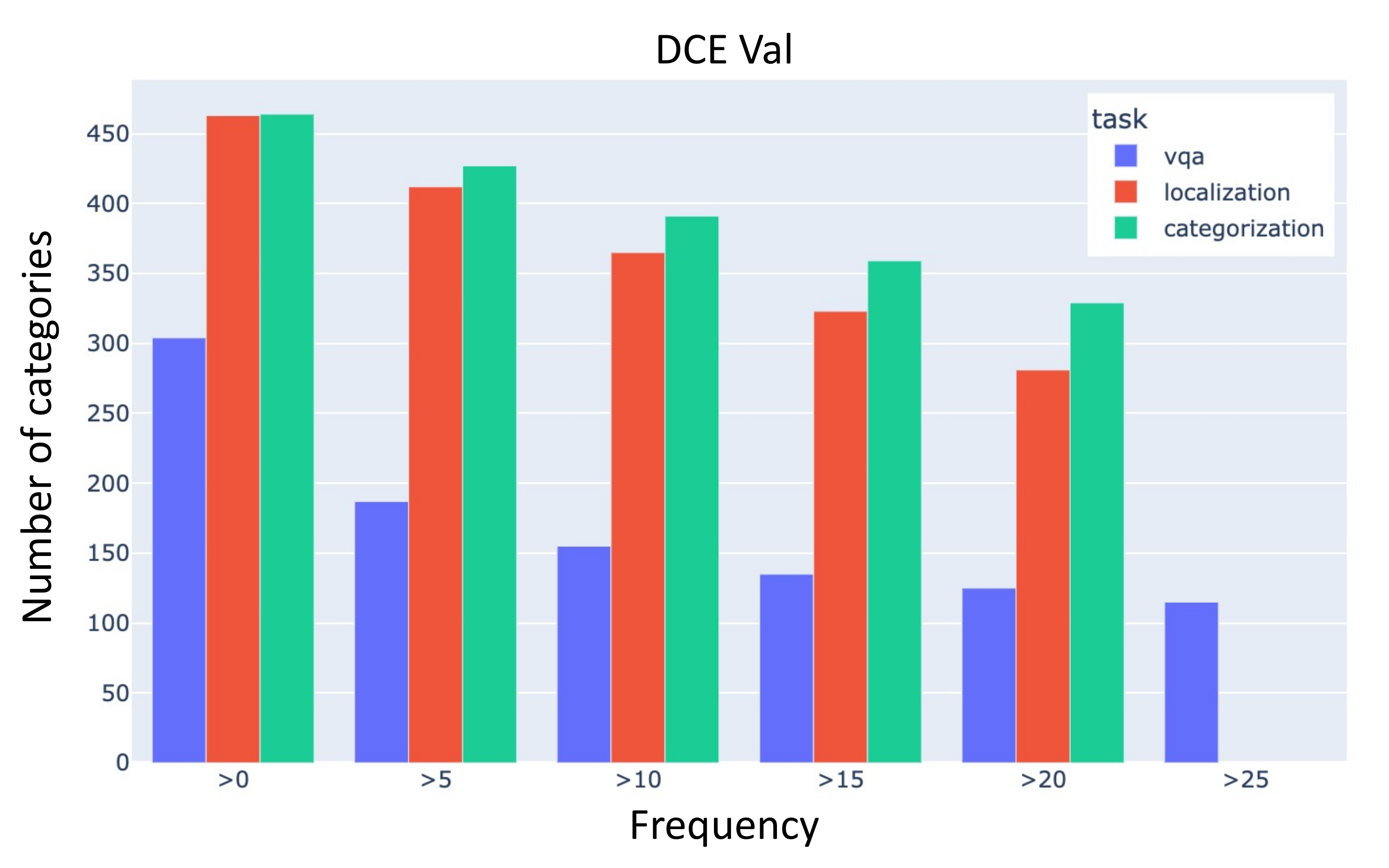}  
  \caption*{}
  \label{fig:dce_freq_val}
\end{subfigure}
\newline
\begin{subfigure}{\linewidth}
  \centering
  % include second image
  \includegraphics[width=0.7\linewidth]{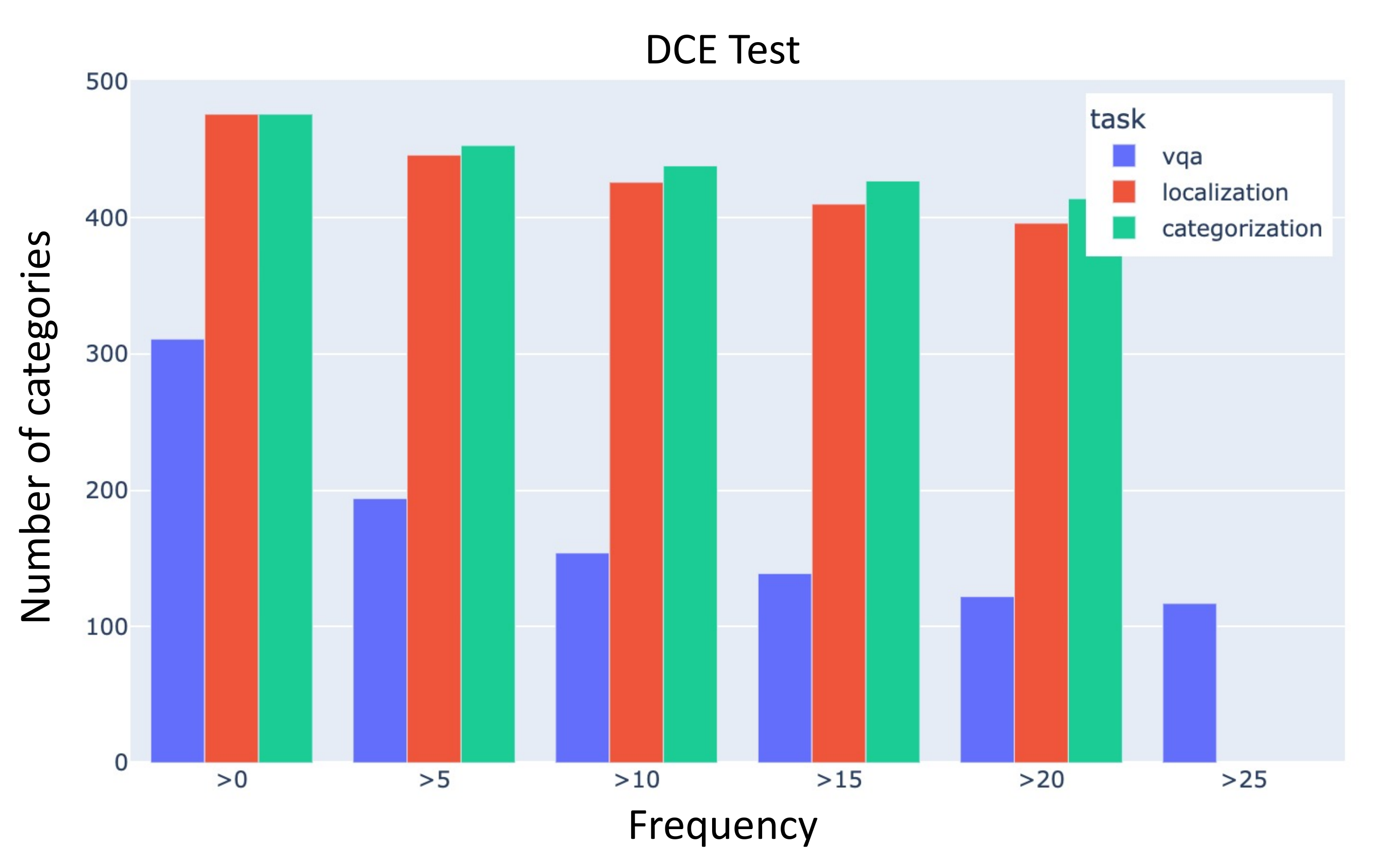}  
  \caption*{}
  \label{fig:dce_freq_test}
\end{subfigure}
\caption{\textbf{\opensce\ val and test set category frequencies.} Bars at $>x$ indicate the number of categories with at least $x$ samples per category for each \opensce\ skill with publicly available annotations. \opensce\ expands the scope of concept evaluation across skills beyond \coco's 80 concepts and maximizes representation of a large subset of mutually exclusive concepts in \openim\ while avoiding over-representation of ``head" concepts (e.g. ``man", ``woman").}
\label{fig:dce_freq}
\end{figure}

Fig.~\ref{fig:dce_freq} shows the number of categories with various frequencies of occurrence in the \opensce\ val and test sets. Since \textsc{nocaps}~\cite{Agrawal2019nocapsNO} annotations are hidden behind an evaluation server, we are unable to provide category counts for captioning. Note that VQA has fewer concepts for higher frequencies than localization and captioning because of a lack of a sufficient number of question-answer annotations that mention many of the \openim\ categories selected for \opensce. 

\textbf{VQA sampling strategy.} Co-occurrence of concepts in questions and answers makes the sampling strategy for VQA more nuanced than the one followed for  Cls, CiC, and Loc. We iterate over the categories selected for \opensce\ and randomly sample up to 50 samples for each category. Unlike Cls/CiC and Loc, each sample in VQA may consist of multiple categories. If $k$ samples have already been sampled for the $i^{th}$ category in the selected category list due to co-occurrence with previous $i-1$ categories, we only sample $\max(0,50-k)$ samples for the $i^{th}$ category. This allows the ``tail" categories from the original dataset to be maximally sampled, while ``head" categories are skipped if already sufficiently represented in the annotations sampled thus far. 

\section{\gpvtwo~efficiency metrics}
\label{app:efficiency}

\begin{table}[]
\footnotesize
    \centering
        \caption{\textbf{Number of parameters and FLOPs in \gpvtwo}. Results are shown for both when the image features are pre-computed (top), and when they have to be generated from scratch (bottom). \\}
    \begin{tabular}{c|c|c|c|c|c|c}
        \toprule
         Pre. & Params & VQA & Cap & Loc & CLS & CiC \\
         \midrule
         \checkmark & 224M & 4.68G & 6.31G & 25.1G & 2.63G & 4.73G \\ 
         - & 370M & 7.35T & 7.38T & 7.64T & 6.62T  & 7.30T \\ 
         \bottomrule
    \end{tabular}
    %\vspace{0.05in} %%
    \label{tab:efficiency-metrics}
\end{table}

We report efficiency metrics on \gpvtwo~when features must be extracted from the input image from scratch using VinVL, and for when those features are assumed to have been precomputed. We report parameter count and the number of floating-point operations (FLOPs). Since the number of FLOPs depends on the length of the input, the length of the target text, and the number of regions in the image, we report the average number of FLOPs needed to process a single example on 100 random examples from the training sets for each task. We compute FLOPs using a pytorch profiler\footnote{\url{https://github.com/facebookresearch/fvcore/blob/main/docs/flop_count.md}} while computing the loss with a single forward pass of the model. Results are shown in Table~\ref{tab:efficiency-metrics}. We find captioning is slow due to the long output sequences, classification is fast because the output text is short and there tends to be fewer objects in the cropped classification images, and detection requires generating per-box outputs so it requires the most compute. If computing the features from scratch, the computational cost is dominated by VinVL, which requires running a X152-FPN backbone and computing features for a large number of proposal regions~\cite{Zhang2021VinVLMV}.

\section{Experimental Details}
\label{app:training_details}
Here we give a more detailed account of how the models are trained. We train \gpvtwo\ and \vltf\ using the Adam optimizer~\cite{adam} with a batch size of 60 and learning rate of 3e-4, $\beta_1$ of 0.9, $\beta_1$ of 0.999, $\epsilon$ of 1e-8, and a weight decay of 1e-4. The learning rate linearly warms up from 0 over 10\% the training steps and then linearly decreases back to 0. The web data is sharded into 4 parts, and a different part of used for each epoch for the first four epochs. Then the data is re-sharded into 4 new parts for the final 4 epochs. The data is stratified so that the 6 supervised datasets (VQA, Cap, Loc, CLS, CiC and the current web shard) are represented in approximately the same proportion in each batch. During training, we use the cross-entropy loss of generating the output text for all tasks besides localization. For localization, we compute relevance scores for each box following the process in Sec.~\ref{sec:model} and then train using the Hungarian-matching loss from DETR~\cite{Carion2020DETR} with two classes (one class for relevant and one for irrelevant) following~\cite{gpv1}. We compute the scores on the in-domain validation sets each epoch, and use the checkpoint with the highest average score across all validation tasks. 
We experimented with using different learning rates for \vltf\ but found it had little impact on performance, so used the same learning rates for both models. We use the prompts created by \cite{gpv1} for CLS, Loc and Cap, and from our questions template for \web~(See Sec.~\ref{app:web-questions}). For CiC we use the CLS prompts.
During testing, we generate text using beam search with a beam size of 20, except for classification on \opensce~in which case we use the ranking approach from Sec.~\ref{sec:model}.

\section{Human Object Interaction experimental details}
\label{app:hoi_details}
In this section, we provide more details about how \gpvtwo~is trained to perform human-object interaction. Both stages of the two-pass process from Sec.~\ref{sec:model} are trained using the HOI-Det training set~\cite{ChaoHOIDet}. The first pass requires the model to locate person bounding boxes in the image, \gpvtwo~is trained to do this by using localization examples constructed from the HOI annotations. In particular, we build examples by gathering all person-boxes in the annotations for an image and then pruning duplicate boxes by applying non-maximum suppression with a threshold of 0.7. The remaining boxes serve as ground truth for localization examples with the prompt \prompt{Locate the people}. 

The second pass requires the model to identify object interactions given a person box. \gpvtwo~is trained using the same de-duplicated person boxes from the HOI annotations. For each such person box, the input to the model is the image with the prompt \prompt{What is this person doing?} and the input query box set to be the person box. Target outputs are built by gathering all HOI annotations for that input person box (annotations with person boxes that were pruned during de-duplication are mapped to the person box with the highest IoU overlap). This results in a set of object boxes labeled with HOI classes for each person box. Those object boxes are aligned with the boxes found by the object detector by finding the box with the highest IoU overlap with each ground truth object box. During training, if no box from the object detector has at least a 0.5 overlap with an object box, we manually add that object box to the regions extracted by the detector so we can still train on it. The model is trained to generate a text description of the HOI class for each box that was aligned with a ground truth box (e.g., ``riding the horse" for the HOI class riding+horse), or the text ``no interaction" for any box that was not aligned with a ground truth object. In practice, we only train on a randomly selected half of the ``no interaction" boxes to reduce computational expense. If an object box is aligned to multiple ground truth boxes, and therefore has multiple HOI class labels, we train the model to generate all such labels with a high probability.

We train the model with the hyper-parameters specified in Sec.~\ref{app:training_details}, but for 4 epochs with a batch of 48 and a learning rate of 1e-4. Since this task is intended as a demonstration, we did not spend a lot of time optimizing this process and think it could be further improved with additional effort.

To evaluate the model, we first find boxes the model identifies from the prompt \prompt{Locate the people} with a score of over 0.5. Then for each such box, for each object box detected by the object detector, and for each HOI class, we score the box pair and class with the log-probability of generating the class label text from the object box when the person box is used as the input query box. 
In practice, for a given person box, we prune object boxes that generate the text ``no interaction" with a high probability so we do not have to score a generation for every class label with that box-pair. 
These scores are finally used to compute the average precision metric from~\cite{ChaoHOIDet}. 

Finding HOIs for an image requires one forward pass with the encoder for each person box, then one forward pass with the decoder for each person box/object box pair to compute the ``no interaction" probability, and then another forward pass with the decoder for each person box, non-pruned object box, and class label to get the class scores. This is made affordable by the fact the class labels are short, and we are able to label the 10k test set in about an hour using a single Quadro RTX 8000 GPU (after the VinVL image features have been precomputed).

% [training/testing] \gpvtwo\ can also be used in a multi-pass procedure to perform niche/complex tasks that are not natively supported in a single pass. Consider the task of Human-Object-Interaction detection~\cite{ChaoHOIBench,ChaoHOIDet} which can be solved via a 2-pass procedure. For a given image, we first locate the person boxes using the text prompt {\em Locate the people} (with the input box set to be the entire image) and using the box-ranking method described above to get the top-n boxes for the word ``person''. Then for each person box, we use the image, the prompt {\em What is this person interacting with?}, and the person box as the input, and compute the log-probability of generating object-interaction phrase options in the dataset for each box found by the object detector (e.g. ``watch the giraffe'')\footnote{We speed this process up by pruning boxes that have a high probability of generating a background phrase, ``no interaction" before computing the probabilities for each object-interaction class.}.

\section{Zero-shot verb and attribute recognition}
\label{app:verbs_attr}

\input{tables/tab_zero_shot}
In addition to nouns, \web\ consists of compositions of nouns with verbs and adjectives. To test the learning of verbs and attributes from \web, we evaluate \gpvtwo\ zero-shot on an action recognition dataset (ImSitu actions~\cite{imsitu}) and an attribute recognition dataset (VAW~\cite{vaw}), see Table~\ref{tab:zero_shot}.
For ImSitu actions we prompt the model with \prompt{What are they doing?}. \gpvtwo\ gets 34.7 top-5 accuracy compared to 58.6 from the benchmark authors~\cite{imsitu} employing a supervised CNN+CRF approach and 68.6 from a recent supervised model\cite{suhail2019mixture} that uses a specialized mixture-kernel attention graph neural network. For verbs present in \web\ (the Seen column), \web\ training provides a significant boost (54.4 from 33.4) showing successful transfer from web images to ImSitu images.
For VAW, we prompt the model with yes/no questions (e.g., \prompt{Is this object pink?}) along with the target object's bounding box to get per-box multi-label attribute results. We see no gains on VAW from \web, likely because the model already learns these attributes from VinVL, \cc, VQA, and Captioning training data.

%Even without \web, 0-shot \gpvtwo\ performs surprisingly well (53.2 vs. 68.3 mAP for a fully supervised model~\cite{vaw}), likely because the model already learns these attributes from VinVL, \cc, VQA, and Captioning training data.

% \section{Specialized Models on~\opensce}
% \label{app:dce_specialized}
% Here we discuss the performance of specialized models on the \opensce\ benchmark. While our work focuses on improving and comparing GPV models, we present these results in order to provide additional calibration of the benchmark.

% \input{tables/tab_dce_benchmark}

\section{Performance on the GRIT benchmark}
\label{app:grit_numbers}
We submit \gpvtwo\ to the Unrestricted track of the GRIT benchmark \cite{Gupta2022GRIT} and achieve state-of-the-art performance at the time of submission. We re-train \gpvtwo\ to include RefCOCO+ \cite{Kazemzadeh14ReferItGame} in the multi-tasking framework in order to compete on the Referring Expressions Grounding task of the benchmark. See Table \ref{tab:grit} for performance results of the model on the test set. 
The results use the \emph{acc.any.agg.$<$task$>$} metric, which averages performance of the model on ``same'' and ``new'' source data for each task, as defined in \cite{Gupta2022GRIT}. Note that \gpvtwo\ is trained on more data than \gpvone, and the VinVL backbone used in \gpvtwo\ is trained on \openim, which belongs to the GRIT ``new'' data source (as allowed by the Unrestricted track), contributing to its performance.

The GRIT benchmark website\footnote{\url{https://grit-benchmark.org/}} contains additional information on the data and the models' ability to generalize to new data sources and concepts, robustness to image distortions, and calibration.

\input{tables/tab_grit}

\section{Comparison between the \gpvtwo\ and \gpvone\ architectures when trained on the same data}
\label{app:direct_comparison}
We now provide an additional comparison between \gpvtwo\ and \gpvone\ in Table \ref{tab:direct_comparison} using the same training data and detector backbone (frozen DETR), trained only on COCO-SCE. This shows that \gpvtwo\ provides gains over \gpvone\ on 3 tasks purely due to its architecture. In addition, adding web data training to \gpvtwo\ (no other changes) provides further improvements on 2 tasks in-domain. Row [c] corresponds to Table 3 in the main paper. 

\input{tables/tab_direct_comparison}

\section{Results on all nocaps splits for \opensce\ captioning}
\label{app:nocaps_full_results}
See Table \ref{tab:nocaps_full} for results of the GPVs on all splits of the nocaps dataset \cite{Agrawal2019nocapsNO}: \emph{in-domain}, \emph{near-domain}, \emph{out-of-domain}, and \emph{all}. The out-of-domain results are reported in the main paper, as our focus is on learning novel concepts.

\input{tables/tab_nocaps_full}

\section{Biases in web data}
\label{app:neg_impact}
We employ several measures to ensure \web~is clean including the ``isFamilyFriendly'' filter on Bing, removing inappropriate words per a popular blacklist~\cite{badwords}, and conducting manual spot checks. 
However, the entire dataset has not been human-curated, so we cannot guarantee it is free from objectionable imagery.
It is important to be aware that search results are known to reflect human biases and stereotypes~\cite{otterbacher2017competent,kay2015unequal}, for example, most of our images for ``soccer player" are of men.
\coco, our main source of supervision, also suffers from these kinds of biases~\cite{zhao-etal-2017-men} so we do not recommend using the models in this paper in production settings.

% \section{License information}
% \label{app:license}
% \noindent Licenses for datasets used in this work are:

% \begin{itemize}
%     \item \ccfull~\cite{Sharma18Conceptual}: A custom open-source license~\footnote{https://github.com/google-research-datasets/conceptual-captions/blob/master/LICENSE}
%     \item \textsc{nocaps}~\cite{Agrawal2019nocapsNO}: Creative Commons Attribution 2.0 Generic
%     \item \vg~\cite{Krishna17Visual}: Creative Commons Attribution 4.0 International License
%     \item \vqa~\cite{goyal2017cvpr_balanced_vqa_v2}: Commons Attribution 4.0 International License
%     \item \coco~\cite{Lin14Microsoft}: Creative Commons Attribution 4.0 License
%     \item \openim~\cite{Kuznetsova18OpenImages}: Creative Commons Attribution 4.0 License
% \end{itemize}

% Licenses for code libraries used in this work are attached separately in licenses.txt. Our code additionally uses elements from the GPV-1 code base\footnote{https://github.com/allenai/gpv}~\cite{gpv1} (Apache 2.0 license). 

%% file: tables/tab_zero_shot.tex
\begin{table}
\setlength{\tabcolsep}{3pt}
\footnotesize
\centering
\caption{
\label{tab:zero_shot}
\textbf{Learning verbs and attributes from \web}. We test verb and attribute learning from \web\ by evaluating \gpvtwo\ without further finetuning on verb (imSitu) and attribute recognition (VAW) benchmarks. \\}
\resizebox{0.7\textwidth}{!}{
\begin{tabular}{
    l
    >{\columncolor{Color5}}c>{\columncolor{Color5}}c>{\columncolor{Color5}}c
    >{\columncolor{Color5b}}c>{\columncolor{Color5b}}c>{\columncolor{Color5b}}c
    }

\toprule

\multicolumn{1}{c}{} 
    & \multicolumn{3}{c}{\cellcolor{Color5}\textbf{imSitu (top-1 $\mid$ top-5 acc.)}}
    & \multicolumn{3}{c}{\cellcolor{Color5b}\textbf{VAW (mAP)}} \\

\multicolumn{1}{c}{Model}
    & \cellcolor{Color5}\emph{Test} & \cellcolor{Color5}\emph{Seen} & \cellcolor{Color5}\emph{Unsn}
    & \cellcolor{Color5b}\emph{Test} & \cellcolor{Color5b}\emph{Seen} & \cellcolor{Color5b}\emph{Unsn} \\
    
\midrule
\gpvtwo  & 10.0 $\mid$ 23.0 & 15.6 $\mid$ 33.4 & 2.5 $\mid$ 9.1 & 53.2 & 56.9 & 52.0 \\
\gpvtwo +web & 16.7 $\mid$ 34.7 & 27.5 $\mid$ 54.4 & 2.2 $\mid$ 8.3 & 52.4 & 56.2 & 51.3 \\
Supervised & 43.2 $\mid$ 68.6 & - & - & 68.3 & - & - \\
\bottomrule
\end{tabular}}
\end{table}

%% file: tables/tab_grit.tex
\begin{table}
\setlength{\tabcolsep}{3pt}
\footnotesize
\centering
\caption{
\label{tab:grit}
\textbf{Performance on GRIT benchmark, unrestricted test set}. \gpvtwo\ competes on four of the seven benchmark tasks: Object Categorization (cat), Object Localization (loc), VQA (vqa) and Referring Expression Grounding (ref). It cannot compete on Segmentation (seg), Person Keypoint Detection (kp), or Surface Normal Estimation (sn). The aggregation takes the average of all seven tasks, assigning $0$ to the tasks that models cannot perform. \gpvone\ here has not been trained on referring expressions, or with web data.\\}
\resizebox{0.8\textwidth}{!}{
\begin{tabular}{
    lc
    >{\columncolor{Color5}}c>{\columncolor{Color5}}c>{\columncolor{Color5}}c
    >{\columncolor{Color5}}c>{\columncolor{Color5}}c>{\columncolor{Color5}}c>{\columncolor{Color5}}c>{\columncolor{Color5b}}c
    }

\toprule

% \multicolumn{1}{c}{} 
%     & \multicolumn{3}{c}{\cellcolor{Color5}\textbf{imSitu (top-1 $\mid$ top-5 acc.)}}
%     & \multicolumn{3}{c}{\cellcolor{Color5b}\textbf{VAW (mAP)}} \\

\multicolumn{1}{c}{Model} & Detector Backbone
    & \cellcolor{Color5}cat & \cellcolor{Color5}loc & \cellcolor{Color5}vqa
    & \cellcolor{Color5}ref & \cellcolor{Color5}seg & \cellcolor{Color5}kp &
    \cellcolor{Color5}sn &
    \cellcolor{Color5b}All\\
    
\midrule
\gpvone & DETR, trained on COCO & 33.2 & 42.7 & 49.8 & 26.8 & - & - & - & 21.8 \\
\gpvtwo & VinVL, trained on COCO, VG, & \textbf{55.1} & \textbf{53.6} & \textbf{63.2} & \textbf{52.1} & - & - & - & \textbf{32.0} \\
 & Objects365 and OpenImages &  &  & &  &  &  &  &  \\
\bottomrule
\end{tabular}}
\end{table}

%% file: tables/tab_direct_comparison.tex
\begin{table}
\centering
\caption{\textbf{Direct comparison between \gpvtwo\ and \gpvone}. Performance on COCO-SCE when trained on the same data and using the same detector backbone.\\}
\label{tab:direct_comparison}
\centering
\scriptsize
\begin{tabular}{
    lcc
    >{\columncolor{Color4}}c
    >{\columncolor{Color4b}}c
    >{\columncolor{Color4}}c
    >{\columncolor{Color4b}}c
    }
\toprule
\multicolumn{1}{c}{} & \multicolumn{1}{c}{Model} & \multicolumn{1}{c}{Web data} 
    & \cellcolor{Color4}\emph{VQA} 
    & \cellcolor{Color4}\emph{Cap} 
    & \cellcolor{Color4}\emph{Loc} 
    & \cellcolor{Color4}\emph{Cls}\\
\midrule
\textcolor{tabindex}{[a]} & \gpvone & no web &  56.4 & 88.3 & \textbf{63.4} & 71.5 \\
\textcolor{tabindex}{[b]} & \gpvtwo & no web & \textbf{59.6} & \textbf{88.4} & 62.2 & \textbf{73.1} \\
\midrule
\textcolor{tabindex}{[c]} & \gpvtwo & with web & 59.9 & 89.2 & 62.2 & 73.0 \\
\bottomrule
\end{tabular}
\end{table}

%% file: tables/tab_nocaps_full.tex
\begin{table}
\centering
\caption{\textbf{Full \opensce\ Captioning results}. Training on web data improves performance for all three GPVs, for all splits --- even in-domain, which focuses on \coco\ concepts. \gpvtwo\ achieves the highest performance by a large margin. \\}
\label{tab:nocaps_full}
\centering
\scriptsize
\begin{tabular}{
    llc
    >{\columncolor{Color2}}c
    >{\columncolor{Color2b}}c
    >{\columncolor{Color2}}c
    >{\columncolor{Color2b}}c
    }
\toprule
\multicolumn{1}{c}{} & \multicolumn{1}{c}{Model} & \multicolumn{1}{c}{Web data} 
    & \cellcolor{Color2}\emph{in} 
    & \cellcolor{Color2}\emph{near} 
    & \cellcolor{Color2}\emph{out} 
    & \cellcolor{Color2}\emph{all}\\
\midrule
\textcolor{tabindex}{[a]} & \gpvone & no web &  69.1 &	51.4 &	25.8 &	49.1 \\

\textcolor{tabindex}{[b]} & \gpvone$^{20}$ & no web &  64.4	& 47.5 &	23.1 &	45.3 \\
\textcolor{tabindex}{[c]} & \gpvone$^{20}$ & with web &  65.7 &	51.2 &	28.6 &	49.0 \\
\midrule
\textcolor{tabindex}{[d]} & \vltf & no web &  70.3 &	55.9 &	31.6 &	53.4 \\
\textcolor{tabindex}{[e]} & \vltf & with web &  72.0 &	60.4 &	45.0 &	59.1 \\
\midrule
\textcolor{tabindex}{[f]} & \gpvtwo & no web & 82.8 &	79.4 &	65.4 &	77.3 \\
\textcolor{tabindex}{[g]} & \gpvtwo & with web & \textbf{85.4} &	\textbf{82.6} &	\textbf{72.5} &	\textbf{81.2} \\
\bottomrule
\end{tabular}
\end{table}